\documentclass[runningheads]{llncs}

\usepackage{eccv}

\usepackage{eccvabbrv}

\usepackage{graphicx}
\usepackage{booktabs}
\usepackage{microtype}
\usepackage{dsfont}

\usepackage{tikz}
\usepackage{ifthen}
\usetikzlibrary{matrix,calc}

\usepackage{bbm}
\usepackage{gensymb}

\usepackage[accsupp]{axessibility}  
\usepackage{algorithm}
\usepackage{algpseudocode}
\usepackage[T1]{fontenc}
\usepackage{float}

\usepackage{pgf-pie} 

\usepackage[pagebackref,breaklinks,colorlinks,citecolor=eccvblue]{hyperref}

\usepackage{orcidlink}

\usetikzlibrary{shapes.geometric, arrows}
\usetikzlibrary{shapes.multipart}
\usetikzlibrary{calc}
\usetikzlibrary{fit, positioning}

\definecolor{datasetGreen}{HTML}{93C47D}
\definecolor{baseBlue}{HTML}{CFE2F3}
\definecolor{componentOrange}{HTML}{F6B26B}

\definecolor{umlYellow}{HTML}{fffdcc}
\definecolor{umlPurple}{HTML}{ccccff}

\tikzset{font=\fontfamily{lmtt}\selectfont}
\tikzstyle{arrow} = [very thick, rounded corners, ->,>=stealth]

\tikzstyle{base} = [
    rectangle,
    rounded corners,
    minimum width=2cm, 
    minimum height=2cm,
    text centered, 
    draw=black, 
    very thick,
    fill=umlPurple
]

\tikzstyle{image} = [
    rectangle,
    draw=black,
    very thick,
    inner sep=0,
    outer sep=0
]

\DeclareMathOperator*{\argmin}{arg\,min}

\begin{document}

\title{PixLens: A Novel Framework for Disentangled Evaluation in Diffusion-Based Image Editing with Object Detection + SAM} 

\titlerunning{PixLens}
\def\thefootnote{*}\footnotetext{These authors contributed equally to this work.}
\author{Stefan Stefanache\inst{1}\thefootnote{*}\and
Lluís Pastor Pérez\inst{1}\thefootnote{*}\and
Julen Costa Watanabe\inst{1}\thefootnote{*} \and
Ernesto Sanchez Tejedor\inst{1}\thefootnote{*} \and
Thomas Hofmann\inst{1} \and
Enis Simsar\inst{1}}

\authorrunning{S. Stefanache et al.}

\institute{ETH Zurich, Rämistrasse 101, 8092 Zürich, Switzerland\\
\email{sstefanache@ethz.ch, lpastor@ethz.ch, jucosta@ethz.ch, sernesto@ethz.ch, thomas.hofmann@inf.ethz.ch, enis.simsar@inf.ethz.ch}}
\maketitle

\begin{abstract}
Evaluating diffusion-based image-editing models is a crucial task in the field of Generative AI. Specifically, it is imperative to assess their capacity to execute diverse editing tasks while preserving the image content and  realism. While recent developments in generative models have opened up previously unheard-of possibilities for image editing, conducting a thorough evaluation of these models remains a challenging and open task. The absence of a standardized evaluation benchmark, primarily due to the inherent need for a post-edit reference image for evaluation, further complicates this issue. Currently, evaluations often rely on established models such as CLIP or require human intervention for a comprehensive understanding of the performance of these image editing models.
Our benchmark, PixLens, provides a comprehensive evaluation of both edit quality and latent representation disentanglement, contributing to the advancement and refinement of existing methodologies in the field.
\keywords{Diffusion-based Image Editing, Generative AI, Evaluation Benchmark, Latent Representation Disentanglement}
\end{abstract}

\section{Introduction}
\label{sec:intro}

Recent developments in Generative AI, particularly in large-scale text-to-image diffusion models \cite{stablediffusion, ho2021cascaded}, have enabled the development of text-guided image editing models \cite{instructpix2pix, controlnet, mokady2022nulltext}. Notably, text-guided image diffusion models \cite{stablediffusion, ho2021cascaded, saharia2022photorealistic} have recently demonstrated substantial image generation capabilities. Pre-trained on extensive image-text pairs like LAION \cite{schuhmann2022laion5b} using a diffusion objective, these models have achieved state-of-the-art FID \cite{FID} scores on benchmarks such as MS-COCO \cite{lin2015microsoft} and exhibited success in editing real images \cite{mokady2022nulltext, Kawar_2023_CVPR}. In the domain of image editing models, recent strides have been made, exemplified by InstructPix2Pix \cite{instructpix2pix}, ControlNet \cite{controlnet}, LCM \cite{lcm}, OpenEdit \cite{openedit}, and VQGAN+CLIP \cite{vqgan}, which leverage established image generators like Stable Diffusion \cite{stablediffusion} and GANs \cite{GAN}.

However, progress is hampered by the absence of a widely accepted and standardized methodology for evaluating the performance of image editing models. The absence of a post-edited image reference makes this task somewhat subjective and non-trivial. Efforts to accomplish this task have already been made in the past, however existing evaluation methods may not be universally applicable or suitable for all situations. The majority of benchmarks overly depend on the CLIP score \cite{clip}, which is not always reliable \cite{goel2022cyclip}, as discussed in Section \ref{sec:magicb}, or on distribution-based metrics such as FID \cite{FID}. While FID is useful for assessing the overall quality and diversity of generated images, it does not account for specific edit accuracy or the preservation of image attributes and has been proven to be biased \cite{FID_biased}, making it even less reliable. This over-reliance on distribution-based metrics and CLIP scores can lead to misleading evaluations, where models appear performant but fail to meet nuanced or task-specific criteria.

In light of this challenge, we propose PixLens, an automatic benchmarking framework designed to address the shortcomings of current evaluation methodologies for diffusion-based image editing models. PixLens emphasizes automatic evaluation and leverages SAM-based detection models \cite{sam,owlvitsam,groundedsam} for precise object localization through segmentation masks. In addition, it explores the correlation between latent space disentanglement and image editing performance, offering a comprehensive approach to model evaluation. Our contributions include:

\begin{itemize}
    \item Automatic evaluation procedures that mitigate subjectivity and provide standardized assessments.
    \item Utilization of advanced detection and segmentation models to enhance the evaluation process, particularly for complex objects and scenarios.
    \item Comprehensive assessment of both subject and background preservation in edited images, including untargeted elements not intended for modification.
    \item Introduction of multiplicity handlers to address issues such as hallucinations in edits and ambiguous scenarios, ensuring robust evaluation outcomes.
    \item Exploration of latent space disentanglement and its influence on image editing performance, providing insights into model behavior.
    \item Streamlined pipeline setup and usage designed for ease-of-use by image editing model developers, facilitating efficient evaluation workflows.
\end{itemize}




\begin{figure}[t!]
\centerline{
    \scalebox{0.475} {
        \input{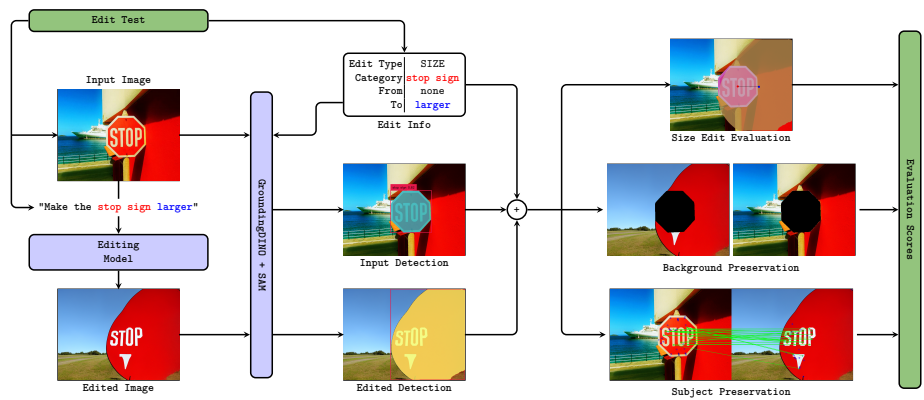}
    }
}
\caption{PixLens Edit Evaluation Pipeline: \texttt{SIZE} operation evaluation example.}
\label{fig:pixlens}
\end{figure}


\section{Related Work}


Benchmarks such as 
PIE-Bench \cite{pie-bench}
and MagicBrush \cite{magicbrush}, present different automated pipelines for evaluating the quality of edits. Yet, a common limitation is their over-reliance on CLIP \cite{clip} scores, which has been shown to be \emph{unreliable} in certain cases, as noted in \cite{goel2022cyclip} and exemplified in Section \ref{sec:magicb}. ImagenHub \cite{ku2023imagenhub} addresses this concern by proposing a unified framework incorporating human evaluation scores, albeit introducing the inconvenience of expert raters. 

EditVAL \cite{basu2023editval} employs OwL-ViT \cite{owlvit} for automated evaluation. However, when facing hallucinations from the image editing model, such as the addition of unintended objects or drastic changes to unintended parts of the image (e.g., background), EditVAL fails to provide a clear rationale for handling these issues. It remains ambiguous how their proposed benchmark addresses situations with multiple objects of the same category in the edited image—whether it considers a single instance or adopts an alternative strategy. Moreover, in scenarios involving complex shapes, relying only on bounding boxes introduces imprecision. 

Other noteworthy text-guided image editing benchmarks include TedBench 
\cite{Kawar_2023_CVPR} and EditBench \cite{editbench}, each with their limitations. TedBench evaluates on a relatively small dataset and lacks assessment for recent popular methods like InstructPix2Pix \cite{instructpix2pix}, while EditBench is limited to mask-guided evaluation, i.e. it requires an additional mask along with the edit prompt. 



\section{PixLens: Benchmark for Text-Guided Image Editing}

This section introduces PixLens, our novel automated benchmark for text-guided image editing. It consists of three primary components: (i) an automated editing quality evaluation procedure, (ii) a subject and background preservation evaluation pipeline, and (iii) a latent space disentanglement analysis.

\subsection{Edit Quality Automatic Evaluation}

\subsubsection{Evaluated Edit Types and Pipeline Structure.}

We draw inspiration from EditVAL's categorization of 13 distinct edit types, although their automatic evaluation pipeline only assesses a subset of 6. Notably, we introduce automatic assessment for \texttt{object-removal}, \texttt{single-instance-object-removal}, and \texttt{color} edit types, which are missing in their benchmark, though we acknowledge limitations regarding color evaluation biases, as discussed in Section \ref{sec:magicb}.

Our benchmark adopts EditVAL's JSON structure for defining edit sets, facilitating easy modification and introduction of new edit types and operations. Specifically, for an edit operation to be evaluated successfully, the following attributes must be defined: \texttt{category}, representing the class name for the most significant object in the image (aligned with the MS COCO dataset class in EditVAL's case); \texttt{from}, describing the original value of the attribute to be changed in the image; and \texttt{to}, indicating the desired attribute value in the edited image. Further details on these attributes and their utilization across different edit types are provided in the Supplementary Material.

This flexible approach allows us to adapt any pre-defined evaluation benchmark dataset to our purposes and execute PixLens seamlessly. We demonstrate this adaptability by applying our pipeline to both EditVAL's dataset (with minimal modification) and MagicBrush's dataset in subsequent sections. This provides qualitative and quantitative comparisons that highlight the alignment of our benchmark with human evaluation criteria.

Our automated evaluation process assigns a score to each edited image, indicating the success or failure for a subset of edit types. Formally, for an original image $\mathbf{I}_{\text{in}}$, an edited image $\mathbf{I}_{\text{edited}}$, an edit type evaluator $\mathcal{E}$, and a possible edit operation $\mathcal{O}$, the per-image edit accuracy is defined as:

$$\text{Edit Accuracy} = \mathcal{E}(\mathbf{I}_{\text{in}}, \mathbf{I}_{\text{edited}}, \mathcal{O}) \in [0, 1]$$

where a score of 1 represents a successful edit (from the desired edit type's perspective), while 0 indicates a failed edit. Our \textbf{9} edit operations $\mathcal{O}$ are: \texttt{Object addition} (incorporating an object into the image), \texttt{Size change} (increasing or decreasing an object's size), \texttt{Positional addition} (placing an object in a specific location), \texttt{Position replacement} (moving an object), \texttt{Object replacement} (substituting one object for another), \texttt{Object removal} and \texttt{Single instance object removal} (removing an object), \texttt{Alter parts} (modifying an object's characteristics), and \texttt{Color change} (changing an object's color). Details on the evaluation rules and methods for each edit type are elaborated in the Supplementary Material.

Unlike EditVAL's binary approach, our scoring system spans the range $[0, 1]$, providing a more nuanced assessment of edit quality. Additionally, we generate a boolean value indicating the success of the evaluation process, which is crucial for discarding evaluations in cases where detection and segmentation processes fail.

To address edge cases and improve evaluation robustness, we introduce ``Multiplicity Handlers''. This approach tackles (i) scenarios involving multiple instances of the targeted category in the input image, and (ii) situations of evaluation ambiguity. For example, in the \texttt{positional-addition} edit, where the objective is to add an object while specifying its position, the final edited image may contain multiple objects added in various locations. Unlike EditVAL, which lacks clarity in handling these scenarios, our ``Multiplicity Handlers'' employ refined logic to address complex situations effectively.

\subsubsection{Detection and Segmentation Integration.}

EditVAL highlights the effectiveness of CLIP \cite{clip} in evaluating the alignment between edited images and the prompts. However, CLIP often falls short in recognizing complex spatial relationships, such as edits where the position or size of the main object in the image is modified, as discussed in \cite{gokhale2023benchmarking}. To address this, EditVAL adopts OwL-ViT \cite{owlvit}, a vision language model known for its fine-grained object localization capabilities.


Expanding upon this approach, we incorporate segmentation masks alongside detection models in our methodology. Segmentation masks offer detailed object boundaries, enhancing our understanding of edits beyond what detection models alone can provide. Drawing inspiration from OwL-ViT+SAM \cite{owlvitsam} and Grounded SAM \cite{groundedsam}, we combine detection and segmentation models to provide label annotations, bounding boxes, and segmentation masks for each object of interest in an image.

This integration not only enables precise area comparisons and assessments of features like SIFT and background preservation but also enhances the accuracy of overlaps and intersections. Compared to EditVAL, our approach involves more sophisticated logic for edit-specific operations, facilitating a more detailed analysis of image edits. 

\subsubsection{Subject Preservation.}

In contrast to existing benchmarks such as EditVAL or MagicBrush, our benchmark uniquely incorporates a comprehensive evaluation of subject preservation. This crucial aspect of image editing quality is often overlooked in related work. While MagicBrush and PIE-Bench use the DINO \cite{dino} metric for subject preservation, we assess the fidelity of the subject across multiple dimensions, providing a more nuanced and intuitive understanding of edit performance.

When providing instructions such as ``Move the ball to the left of the image", we expect that the edited image will retain the same subject as the original input. Based on this concept, we assess subject preservation from four distinct perspectives:

\begin{itemize}
    \item SSIM \cite{SSIM} \& SIFT \cite{SIFT}: Do the structural features remain consistent between the original and edited images?
    \item Aligned IoU: Does the edited image maintain the same size and shape as the original?
    \item Color Similarity: Is the color of the subject preserved?
    \item Normalized Euclidean Distance: Does the edited subject retain its original position?
\end{itemize}

Examples illustrating strong and weak subject preservation are showcased in Figures \ref{fig:good_sp} and \ref{fig:bad_sp}, respectively. In these cases, color preservation scores are disregarded as both edits entail color modification. On one hand, the robust preservation of the \emph{dog} in Figure \ref{fig:good_sp} is characterized by a high ratio of matching SIFT keypoints (0.1), a very high IoU of aligned masks (0.968), good SSIM (0.836), and a minimal normalized distance between the mask centroids (0.0007). On the other hand, the preservation of the \emph{backpack} in Figure \ref{fig:bad_sp} exhibits significant structural changes (SIFT: 0.022, SSIM: 0.588), with alterations in shape and position (Aligned IoU: 0.489, Centroid Normalized Distance: 0.43). Additional details regarding subject preservation evaluation and score computation can be found in the Supplementary Material.



\subsubsection{Background Preservation.}
There remains a lack of consensus regarding the most effective method for evaluating background preservation in masked images. PIE-Bench tackles this issue by employing a variety of metrics such as SSIM, MSE, PSNR and LPIPS \cite{lpips}. Through experimentation with such techniques, we discovered that employing Mean Squared Error calculations between masked images and their grayscale counterparts yielded the most reliable results. This approach effectively accounted for not only color, but also shape and texture differences, providing a comprehensive assessment of background preservation.

\begin{figure}
  \centering
  \begin{subfigure}{0.27\textwidth}
    \centering
    \includegraphics[width=\linewidth]{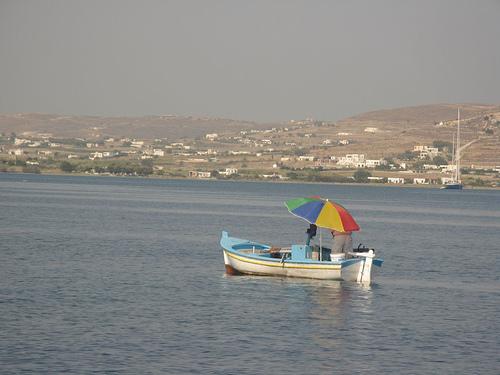}
    \caption{Original image}
  \end{subfigure}
  \begin{subfigure}{0.27\textwidth}
    \centering
    \includegraphics[width=\linewidth]{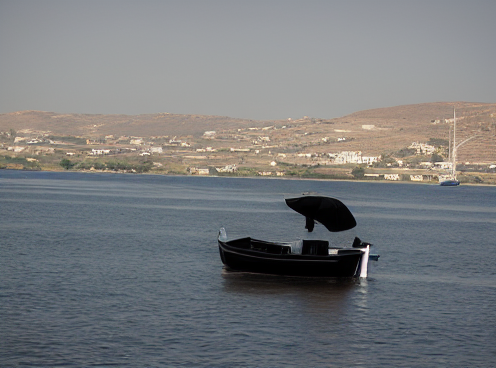}
    \caption{``To black``}
  \end{subfigure}
  \begin{subfigure}{0.27\textwidth}
    \centering
    \includegraphics[width=\linewidth]{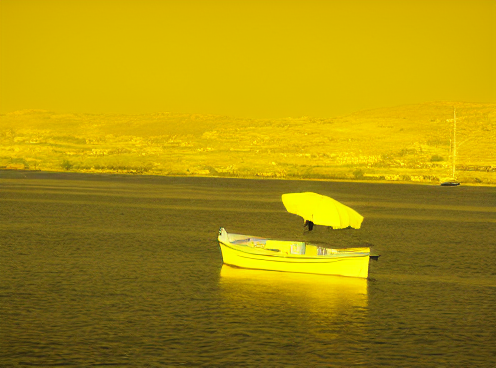}
    \caption{``To yellow''}
  \end{subfigure}
  \caption{Original image (left) and edited images resulting of ``changing the color of the boat to black (center) / yellow (right)'', using InstructPix2Pix.}
  \label{fig:boat_background}
\end{figure}

\begin{figure}
  \centering
  \begin{subfigure}[t]{0.5\textwidth}
    \centering
    \includegraphics[width=\linewidth]{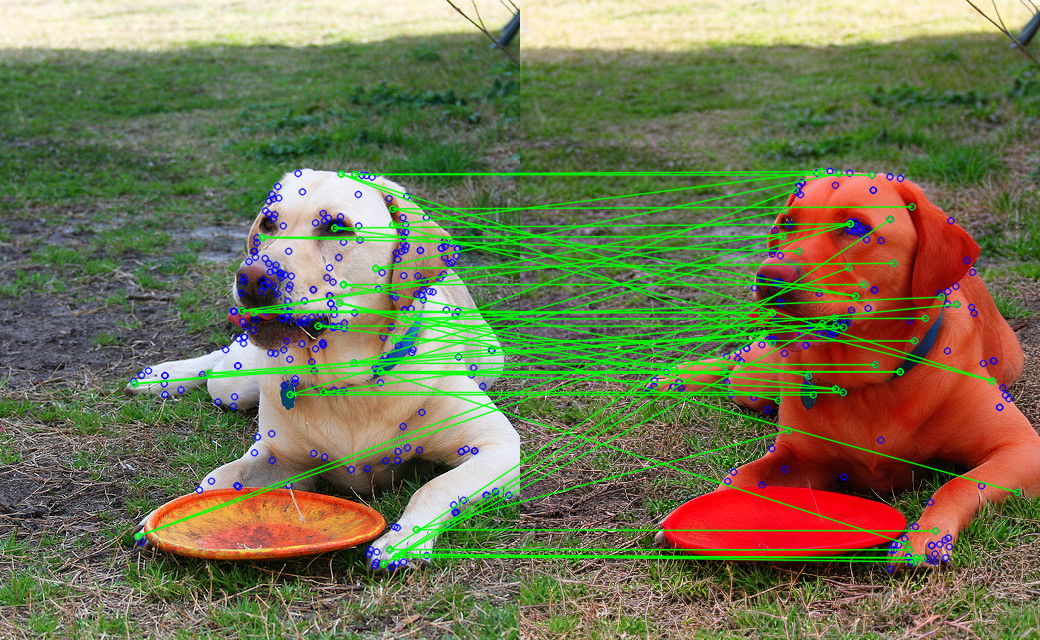}
    \caption{Matching SIFT keypoints for \emph{dog} between the original and edited image}
  \end{subfigure}
  \hspace{2em}
  \begin{subfigure}[t]{0.25\textwidth}
    \centering
    \includegraphics[width=\linewidth]{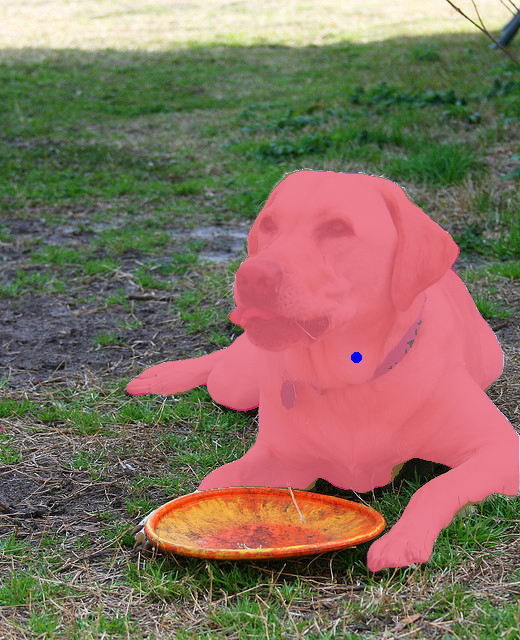}
    \caption{Overlapped masks of the \emph{dog} from input and edited images}
  \end{subfigure}
  \hfill
  \caption{Good subject preservation example for InstructPix2Pix edit with the prompt \emph{Change the color of the dog to red}. The subject of this edit is the \emph{dog}.}
  \label{fig:good_sp}
\end{figure}

\begin{figure}
  \centering
  \begin{subfigure}[t]{0.50\textwidth}
    \centering
    \includegraphics[width=\linewidth]{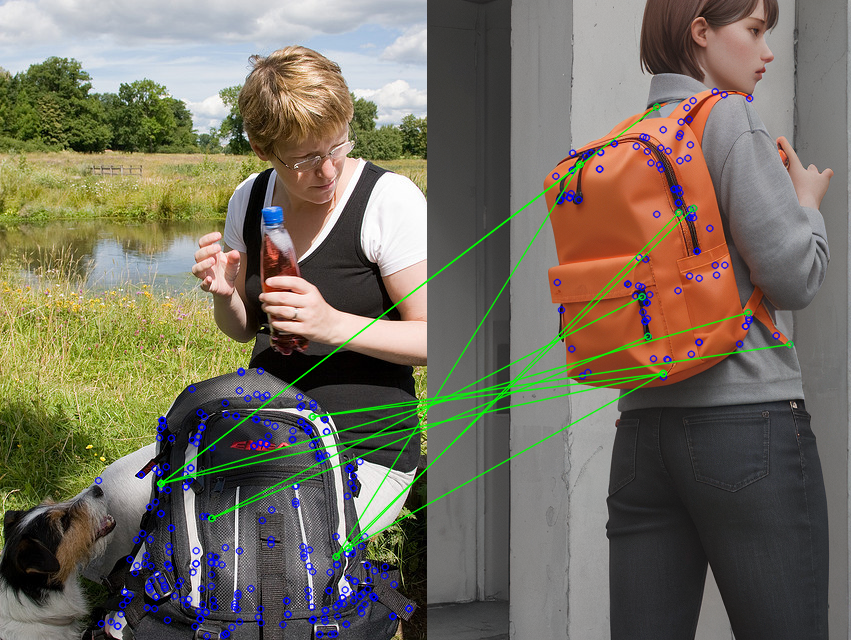}
    \caption{Matching SIFT keypoints for \emph{backpack} between the original and edited image}
  \end{subfigure}
  \hspace{2em}
  \begin{subfigure}[t]{0.25\textwidth}
    \centering
    \includegraphics[width=\linewidth]{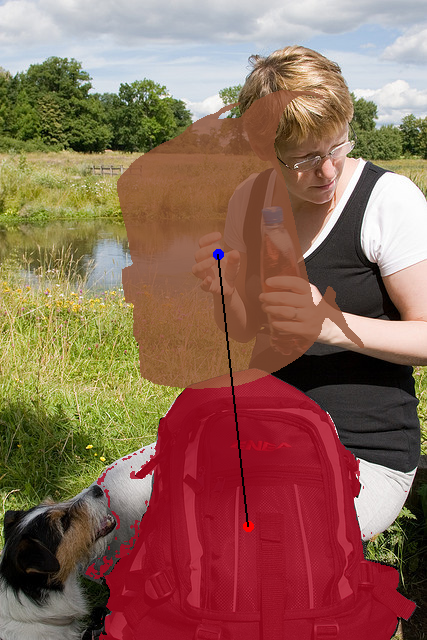}
    \caption{Overlapping masks of the \emph{backpack} from the input and edited images}
  \end{subfigure}
  \hfill
  \caption{Bad subject preservation example for LCM edit with the prompt \emph{Change the color of the backpack to orange}. The subject of this edit is the \emph{backpack}.}
  \label{fig:bad_sp}
\end{figure}

In Figures \ref{fig:boat_background}, \ref{fig:good_sp}, and \ref{fig:bad_sp}, we highlight the significance of evaluating subject and background preservation in image edits, a key feature of our benchmark distinguishing it from others like TedBench or EditVAL. 

In Figure \ref{fig:boat_background}, we compare two \texttt{color} edits using InstructPix2Pix to alter the boat's color in a lake. Despite both edits being successful (both scored 0.99), changing the boat to yellow also affects the background, resulting in preservation scores of 0.89 and 0.40, respectively. This underscores the importance of assessing how well models disentangle various image attributes beyond specific tasks, aiding developers in refining their editing models.

\subsection{Disentanglement Evaluation}
We address a significant gap in current image editing benchmarks: the evaluation of disentanglement.
Disentanglement refers to the process of identifying and separating different factors that make up a complex data representation. Modularity \cite{disentanglement_review} is an important concept that states that various factors in a representation space should operate independently and only affect specific subspaces without interacting with each other. Compactness \cite{disentanglement_review} is another crucial aspect, which suggests that each factor's influence should be limited to as small a subset of the representation space as possible, ideally confined to a single dimension for each factor. This ensures that the representation is clear and precise. Lastly, explicitness \cite{disentanglement_explicitness} refers to the clarity and simplicity of the mapping between the representation space, i.e., the code, and the value of a factor, ideally being a straightforward linear relationship. In this framework, we evaluate modularity and explicitness.
Previous disentanglement studies include the framework proposed in \cite{disentanglement_framework}, where measures of disentanglement, completeness, and informativeness \cite{disentanglement_review} can be obtained, but the ground truth latent structure must be provided, which often is not accessible. Another previous study by \cite{studySDdisentanglement} shows that if the noise parameters are fixed during the denoising process in Stable-Diffusion-based models, there is a way to create disentangled images that surpass a previous baseline.
\subsubsection{Setup and Disentanglement Pipeline.}

We start from a blank white image, crucial in circumventing positional-encoding issues that typically arise in the embedding process. We define a comprehensive list of objects and corresponding attributes to facilitate our evaluation. This includes a diverse range of elements as illustrated in the Supplementary Material.
For each attribute category (e.g., texture, color, pattern), we generate pairs from the respective attribute lists. We denote \(z_{\text{prompt}}\) as the latent representation of an edited white image with a specific prompt. For each attribute pair (\( a_1, a_2\)) in a category, we proceed as follows. First, we generate latent representations \( z_{a_1} \) and \( z_{a_2} \) for the prompts \(a_1\) and \(a_2\). Then, for each object \( o \) in our list, we generate samples with \( z_{\text{start}} = z_{a_1 o} \) and \( z_{\text{end}} = z_{a_2 o} \) (e.g., if \textit{o} is ``chair'' and \( a_1\) is ``blue'', then the prompt would be ``blue chair''). This procedure results in a dataset comprised of tuples of latent encodings of the following form: (\(z_{\text{start}}, z_{a_1}, z_{a_2}, z_{\text{end}}) \).

\subsubsection{Computing Scores.}

We employ different metrics to evaluate disentanglement, depending on whether we evaluate isolated samples, samples with the same attributes, or all samples.

\textbf{Intra-sample Score:}\label{sec:intra-sample}
As mentioned in \cite{disentanglement_explicitness}, an ideal disentanglement representation should be defined by a simple (linear) function from the code to the factors. A method to verify this is by checking if replacing attribute $a_1$ with attribute $a_2$ in the object $a_1 o$ results in the same latent representation as $a_2 o$. We evaluate this linearization by assessing the isolated behavior of samples, evaluating the distance between $ z_{\text{end}}$ and $(z_{\text{start}} + z_{a_2} - z_{a_1})$
and then averaging across all samples. We normalize these values by dividing them by the Euclidean norm of $z_{\text{end}}$, denoted as $||z_{\text{end}}||_2$, to facilitate a standardized comparison.
\begin{figure}
    \centering
    \includegraphics[width=0.65\linewidth]{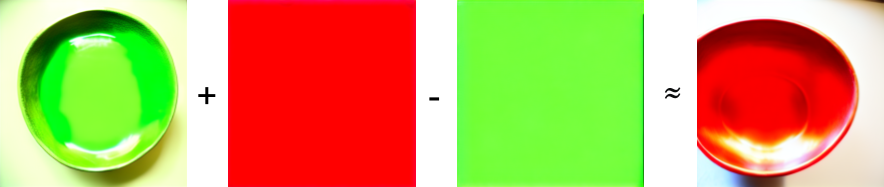}
    \caption{Intra-attribute disentanglement example}
    \label{fig:your_label}
\end{figure}

\textbf{Inter-sample and Intra-attribute Score:}
This part of the evaluation is based on the principle that going from one attribute \( a_{1} \) to another \( a_{2} \) belonging to the same attribute category should follow a given direction independently of the object used. Therefore, we compute vectors \( z_{\text{end}} - z_{\text{start}} \) for every object. We then calculate the average squared cosine similarity across all such vectors.

\textbf{Inter-attribute Score:} 
Finally, based on \cite{betavae,disentanglement_review} we apply a similar procedure to \textit{Z-diff}. The goal is to evaluate the accuracy of a linear model trained to classify differences among samples that share a common attribute. We create a new dataset with entries of the form \(|z_{a o} - z_{a o'}|\) (element-wise absolute difference) for the same attribute $a$, labeled with their attribute type. A linear classifier is then trained to predict the changed attribute. In a disentangled representation, the dimensions corresponding to the same attribute should be close to zero when we do the element-wise absolute difference, which should lead to good performance of the classifier.

\section{Automated Evaluation of State-of-the-Art Models}

\subsection{Performance Overview and Analysis}


We conducted a comprehensive evaluation of five recently introduced text-guided image editing methods using the EditVAL dataset's edit operation collection set. This dataset is particularly well-suited for object-oriented edits, making it an ideal choice for our benchmark's evaluation criteria. The evaluated methods include InstructPix2Pix, ControlNet, Latency Consistency Models, OpenEdit, and VQGAN+CLIP. Our automated evaluation uses SAM on top of Grounding DINO \cite{groundingdino} for object detection and segmentation, with a bounding box extraction confidence threshold set at 0.1.

\begin{table}[h]
    \caption{Edit-specific scores of the considered models for each edit type}
    \label{tab:edit_type_scores}
    \vskip 0.1in
    \begin{center}
    \begin{scriptsize} 
    \begin{sc}
    \begin{tabular}{l@{\hspace{0.5em}}c@{\hspace{0.5em}}c@{\hspace{0.5em}}c@{\hspace{0.5em}}c@{\hspace{0.5em}}c@{\hspace{0.5em}}r} 
    \toprule
    Edit Type & Ins.Pix2Pix & ControlNet & LCM & OpenEdit & VQGAN+Clip & Avg. \\
    \midrule
    Alter Parts    & 0.7143 & 0.7619 & \textbf{0.9286} & 0.7143 & 0.8571 & 0.7952 \\
    Color & \textbf{0.6419} & 0.2539 &  0.1302 & 0.2162 & 0.6367 & 0.3782 \\
    Object Addition & 0.9444 & 0.6667 & \textbf{0.9667} & 0.3889 & 0.9111 &  0.7756 \\
    Object Removal   & \textbf{0.5133} & 0.0471 & 0.0953 & 0.1762 & 0.0639 & 0.1792 \\
    Object Replacement & \textbf{0.9615} & 0.6154 & 0.6923 & 0.5769 & 0.9614 & 0.7615 \\
    Position Replacement & 0.0905 & 0.0000 & \textbf{0.2969} & 0.0372 & 0.1463 & 0.1142 \\
    Positional Addition & 0.2791 & 0.2371 & 0.3531 & 0.0813 & \textbf{0.3655} & 0.2632 \\
    Size   & 0.2600 & 0.1400 & 0.2800 & 0.1500 &\textbf{0.4200} & 0.2500 \\
    \bottomrule
     Avg. & 0.5432 & 0.2969 & 0.4175 & 0.2315 & 0.4895 & \\
    \end{tabular}
    \end{sc}
    \end{scriptsize}
    \end{center}
    \vskip -0.1in
\end{table}

From the results shown in table \ref{tab:edit_type_scores}, we observe that, for simple object manipulation edit types, the scores related to the edit success typically fall within the interval of $[0.25, 0.96]$, with InstructPix2Pix, LCM and VQGAN+CLIP being the top-performing models. On the other hand, in edit types related to spatial manipulation, all models exhibit poor performance, with the potential exception of the LCM and VQGAN+CLIP models, which demonstrate a relatively consistent score. As illustrated in Table \ref{tab:overall_scores}, LCM is one of the worst models in terms of both subject and background preservation, given that its edits naturally deviate significantly from the original image. We interpret this discrepancy as being due to the fact that LCMs" were not originally designed for text-guided image editing, but rather for text-to-image generation, which would allow them to not strictly adhere to the original image, and relocate objects more accurately according to the specified prompt.
\begin{table}[h]
    \caption{Mean edit specific score, subject and background preservation scores of considered models}
    \label{tab:overall_scores}
    \vskip 0.1in
    \begin{center}
    \begin{scriptsize}
    \begin{sc}
    \begin{tabular}{l@{\hspace{0.5em}}c@{\hspace{0.5em}}c@{\hspace{0.5em}}c@{\hspace{0.5em}}c@{\hspace{0.5em}}r}
    \toprule
    Score & Ins.Pix2Pix & ControlNet & LCM & OpenEdit & VQGAN+Clip \\
    \midrule
    Edit-Specific $\uparrow$ & \textbf{0.5432} & 0.2969 &  0.4175 & 0.2315 & 0.4895 \\
    \midrule
    \multicolumn{5}{c}{Subject-Preservation Scores Evaluation} \\
    \midrule
    SIFT $\uparrow$ & \textbf{0.0798} & 0.0167 &  0.0141 & 0.0114 & 0.0063 \\
    Color Preservation $\uparrow$ & \textbf{0.6319} & 0.2449 & 0.4492 & 0.2930 & 0.4550 \\
    Position Preservation $\downarrow$ & 0.0619 & \textbf{0.0426} & 0.1854 & 0.0467 & 0.0956 \\
    Aligned IoU $\uparrow$ & 0.6447 & 0.6852 & 0.3053 & \textbf{0.7103} & 0.4078 \\
    \midrule
    \multicolumn{5}{c}{Background-Preservation Score Evaluation} \\
    \midrule
    Background Preservation $\uparrow$ & \textbf{0.6107} & 0.1491 & 0.2767 & 0.4122 & 0.1996 \\
    \bottomrule
    \end{tabular}
    \end{sc}
    \end{scriptsize}
    \end{center}
    \vskip -0.1in
\end{table}


\subsection{Disentanglement Analysis and Correlation}

\begin{table}
    \caption{Disentanglement scores of considered models}
    \label{tab:disentanglement_scores}
    \vskip 0.1in
    \begin{center}
    \begin{scriptsize}
    \begin{sc}
    \begin{tabular}{l@{\hspace{0.5em}}c@{\hspace{0.5em}}c@{\hspace{0.5em}}c@{\hspace{0.5em}}r}
    \toprule
    Score & Ins.Pix2Pix & ControlNet & LCM & VQGAN+Clip\\
    \midrule
    \multicolumn{5}{c}{Intra-sample Evaluation $\downarrow$} \\
    \midrule
    Avg norm & 1.37 & 1.96 & \textbf{0.89} & 1.84\\
    Avg norm Texture & 1.45 & 1.91 & \textbf{0.84} & 1.85 \\
    Avg norm Color & 1.19 & 1.89 & \textbf{0.89} & 1.82\\
    Avg norm Style & 1.38 & 1.97 & \textbf{0.94} & 1.86\\
    Avg norm Pattern & 1.42 & 2.01 & \textbf{0.87} & 1.85\\
    \midrule
    \multicolumn{5}{c}{Inter-sample Intra-attribute Evaluation $\uparrow$} \\
    \midrule
    Cosine Similarity Texture &\textbf{0.15}& 0.01&0.04 & 0.03 \\
    Cosine Similarity Color &\textbf{0.44}& 0.10&0.07 & 0.09  \\
    Cosine Similarity Style &\textbf{0.09}& 0.01&0.02 & 0.06  \\
    Cosine Similarity Pattern &\textbf{0.18}& 0.01&0.04 & 0.05 \\
    \midrule
    \multicolumn{5}{c}{Inter-attribute Evaluation $\uparrow$} \\
    \midrule
    Accuracy &0.92&0.89&\textbf{0.93}&0.80 \\
    Balanced accuracy &0.92&0.89&\textbf{0.93}&0.83\\
    \bottomrule
    \end{tabular}
    \end{sc}
    \end{scriptsize}
    \end{center}
    \vskip -0.1in
\end{table}

As illustrated in Table \ref{tab:disentanglement_scores} disentanglement evaluation, we can see that overall, LCM and InstructPix2Pix perform better in the three assessments, both of them being Stable-Diffusion-based. We can see that LCM is more linearizable from the intra-sample section, but it does not maintain the attribute directions at the level of InstructPix2Pix. The distinction could arise from the fact that InstructPix2Pix is designed for image editing, so its ability to accurately follow specific instructions is tuned. Conversely, LCM focuses on generating the best possible image, emphasizing retaining the original image structure. In the inter-attribute evaluation, the first three models are quite close, with LCM and InstructPix2Pix being the best-performing ones, but VQGAN+CLIP performs worse, which could also be due to the fact that its latent space is six times greater than the rest. Finally, we can see that the results related to ControlNet in the disentanglement section reflect its performance in the edit-specific scores.


\section{Quantitative Analysis of the Benchmark}

In this section we will present quantitative comparison of our Benchmark with state of the art methods \cite{magicbrush, editbench}, emphasizing how PixLens overcomes the limitation of these benchmarks. 

\subsection{Comparative Analysis with EditVAL} \label{sec:editval}

Our comparison with EditVAL, illustrated in Table \ref{tab:comp_scores}, using InstructPix2Pix as a common model evaluated in both benchmarks, demonstrates that PixLens' computed scores exhibit a notably stronger correlation with human assessment than EditVAL's scores. Specifically, when considering three types of score assignments (EditVAL-human, EditVAL-automatic, and PixLens), our aggregated automatic results show a higher correlation (\textbf{0.903}) with the aggregated human-based scores of EditVAL compared to their automatic method (\textbf{0.146}). 

\begin{table}
    \caption{Edit-specific scores for InstructPix2Pix in EditVAL \cite{basu2023editval} and PixLens}
    \label{tab:comp_scores}
    \vskip 0.1in
    \begin{center}
    \begin{scriptsize} 
    \begin{sc}
    \begin{tabular}{l@{\hspace{2em}}c@{\hspace{2em}}c@{\hspace{2em}}c@{\hspace{2em}}r} 
    \toprule
    Edit Specific Score & PixLens & EditVAL Auto & EditVAL Human \\
    \midrule
    Object Addition & 0.94 & 0.38 & 0.80\\
    Object Replacement & 0.96 & 0.39 &  0.80\\
    Position Replacement & 0.09 & 0.07 & 0.06 \\
    Positional Addition & 0.28 & 0.25 & 0.27\\
    Size   & 0.26 & 0.51 & 0.06 \\
    Alter Parts    & 0.71 & 0.25 & 1.00\\
    \bottomrule
    \end{tabular}
    \end{sc}
    \end{scriptsize}
    \end{center}
    \vskip -0.1in
\end{table}

\begin{figure}
    \centering
    \begin{subfigure}[b]{0.22\textwidth}
        \includegraphics[width=\textwidth]{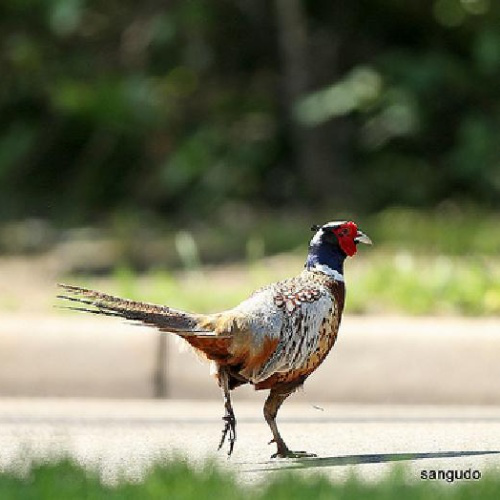}
        \caption{Original}
    \end{subfigure}
    \hfill
    \begin{subfigure}[b]{0.22\textwidth}
        \includegraphics[width=\textwidth]{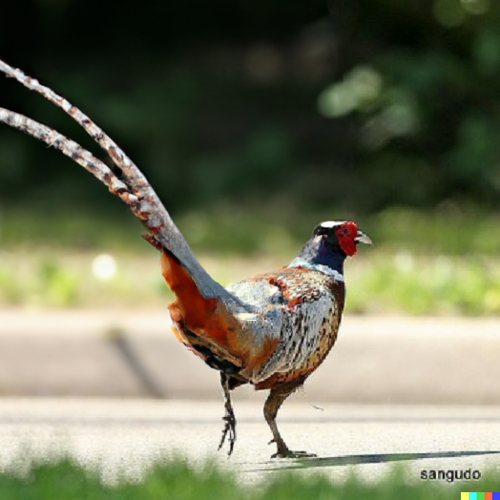}
        \caption{Ground truth}
    \end{subfigure}
    \begin{subfigure}[b]{0.22\textwidth}
        \includegraphics[width=\textwidth]{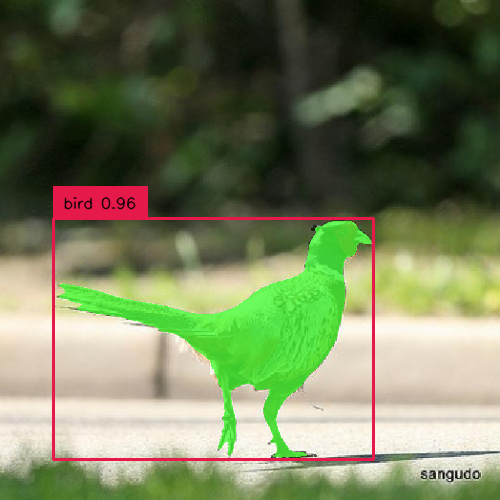}
        \caption{``Bird''}
    \end{subfigure}
    \begin{subfigure}[b]{0.22\textwidth}
        \includegraphics[width=\textwidth]{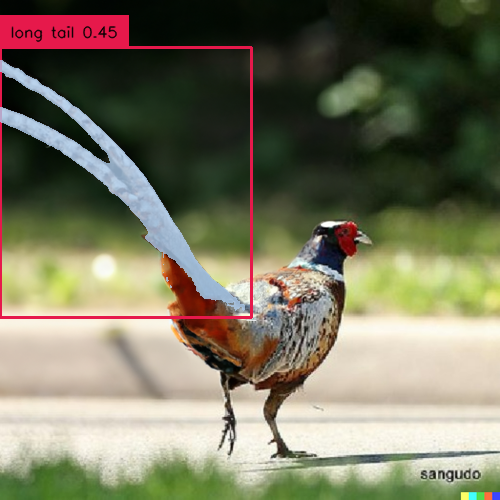}
        \caption{``Long tail''}
    \end{subfigure}
    \caption{MagicBrush instance with prompt ``Give the bird a long tail''. Serves as an example for an \texttt{alter-parts} edit. The tail's bounding box clearly exceeds the bird's bounding box, wrongly scored as a failure when using EditVAL's criteria.}
    \label{fig:alter_parts_example}
\end{figure}

Our enhanced methodology accounts for most of the observed gains in certain operations. For instance, notable improvements in terms of alignment with human evaluation are observed in \texttt{size} operations due to the superior precision of segmentation masks in tracking object areas compared to bounding boxes. 
Additionally, we not only verify whether the size of the object has been altered correctly but also assess whether the relative position of the object remains consistent.
Similarly, significant enhancements are observed in the \texttt{alter-parts} operation, possibly attributed to EditVAL's reliance on containment checks for parts within the larger object, which can be problematic with bounding boxes, particularly when parts are added to the object's boundaries, resulting in reduced scores. Moreover, EditVAL's limited logic assumes that the ``altered part'' cannot exceed the bounds of the modified object, which is often incorrect, as illustrated in Figure \ref{fig:alter_parts_example}. In this example, the tail of the bird is accurately lengthened but extends beyond the bird bounding box's limits. According to EditVAL's criteria, this edit would be erroneously scored as a failure, whereas in our evaluation, we correctly score it as a success.

\subsection{Benchmarking Against MagicBrush Dataset} \label{sec:magicb}

%

To compare the proposed benchmark with MagicBrush, we analyzed all the editing instructions in their ``dev'' file, consisting of 528 edits (the distribution of edit types is shown in the Supplementary Material). We then adapted these instructions to be compatible with our benchmark, assigning an edit type label to each and including all necessary information for evaluation, such as the \texttt{category}, \texttt{from}, and \texttt{to} attributes.  In Table \ref{tab:magicbrush_scores}, we present aggregated scores per edit type obtained from evaluating the ground truth images using PixLens. Overall, the edit-specific scores are generally high, as expected since we are assessing images that were manually altered to serve as ground truth examples. Particularly, edit types such as \texttt{object-replacement} and \texttt{object-addition} demonstrate high scores, reflecting their prevalence in the dataset. However, lower scores are observed for \texttt{object-removal} and \texttt{color} edits. In the case of \texttt{object-removal} edits, the presence of false positives from detection models can lead to misidentification of objects, resulting in lower scores despite successful removal. Regarding \texttt{color} edits, our evaluation methodology exhibits bias towards primary colors, impacting the assessment of color-related edits. This limitation arises due to the subjective nature of color evaluation in image editing, which lacks consensus within the field. This evaluation was a critical step in ensuring that the image quality assessment of PixLens meets human evaluation standards, as the ground truth images used in this evaluation were carefully curated by MagicBrush experts. The evaluation provided a valuable sanity check that helped us verify our alignment with these standards.

\begin{table}
    \caption{Average scores given to MagicBrush's ground-truth images, grouped by edit type}
    \label{tab:magicbrush_scores}
    \vskip 0.1in
    \begin{center}
    \begin{scriptsize} 
    \begin{sc}
    \begin{tabular}{lcccccc} 
    \toprule
     & Edit Specific & SIFT & Color & Position & Aligned IoU & Background \\
    \midrule
    Color Change & 0.38 & 0.12 & - & 0.05 & 0.61 & 0.95\\
    Size Change & 1.0 & 0.02 & 0.12 & 0.17 & 0.15 & 0.98\\
    Object Replacement & 0.89 & - & - & - & - & 0.97\\
    Alter Parts   & 0.65 & - & - & - & - & 0.98\\
    Object Addition & 1.0 & 0.60 & 0.92 & 0.03 & 0.75 & 0.95\\
    Single Instance Removal & 0.5 & - & - & - & - & 0.97\\
    Object Removal & 0.43 & - & - & - & - & 0.95\\
    Positional Addition & 0.73 & - & - & - & - & 0.96\\
    \bottomrule
    Average & 0.73 & 0.37 & 0.89 & 0.04 & 0.69 & 0.96\\
    \bottomrule
    \end{tabular}
    \end{sc}
    \end{scriptsize}
    \end{center}
    \vskip -0.1in
\end{table}

We continue by highlighting critical considerations that distinguish our proposed image quality assessment method from MagicBrush.

\begin{figure}
  \centering
  \begin{subfigure}{0.28\textwidth}
    \centering
    \includegraphics[width=\linewidth]{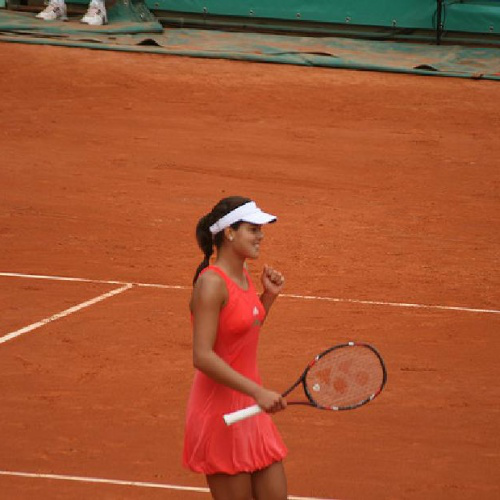}
    \caption*{Original \\ \textbf{``Make her outfit black''}}
  \end{subfigure}
  \begin{subfigure}{0.28\textwidth}
    \centering
    \includegraphics[width=\linewidth]{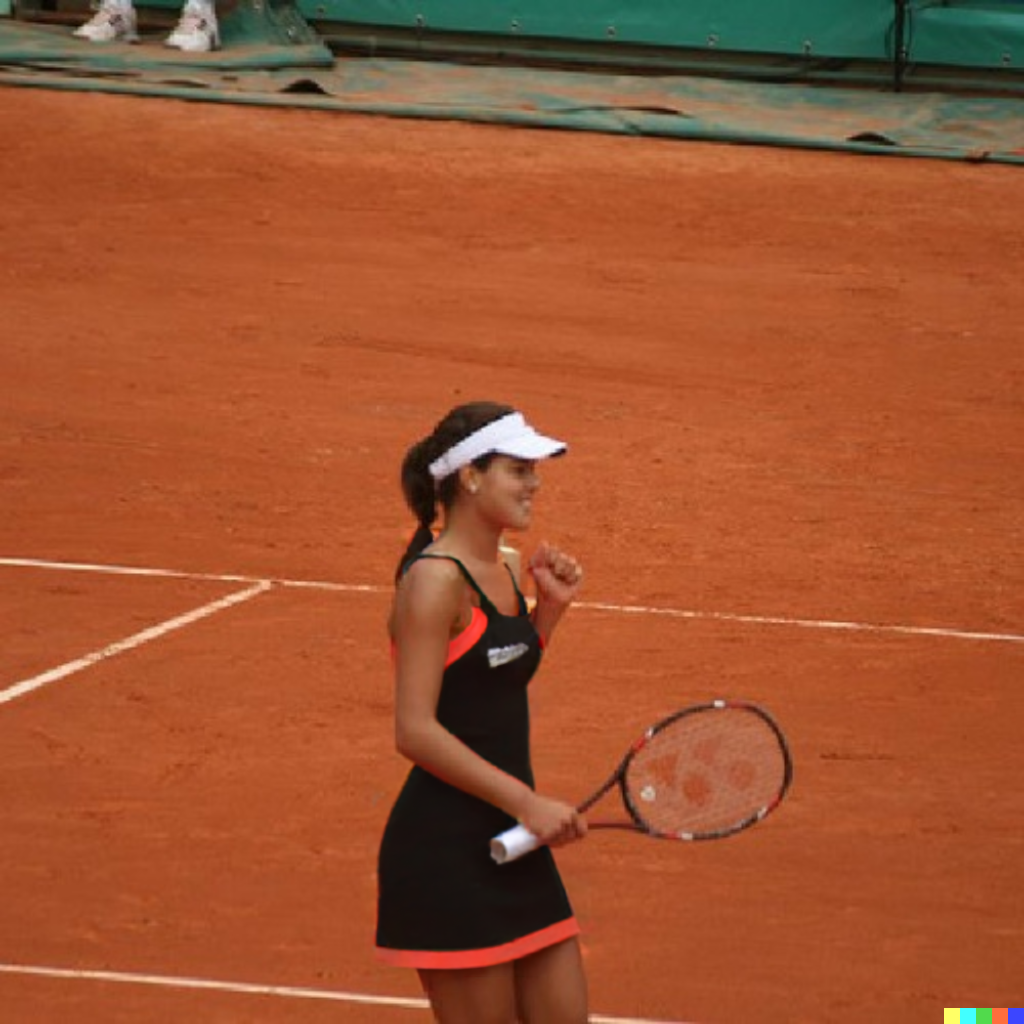}
    \caption*{Ground Truth \\ \textbf{CLIP-T: 0.34}}
  \end{subfigure}
  \begin{subfigure}{0.28\textwidth}
    \centering
    \includegraphics[width=\linewidth]{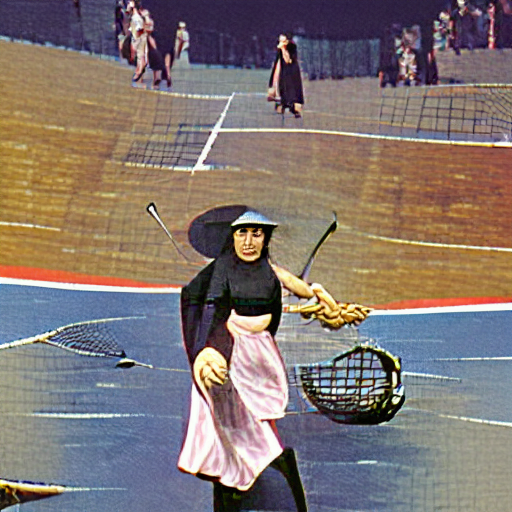}
    \caption*{VQGAN-CLIP \\ \textbf{CLIP-T: 0.55}}
  \end{subfigure}
  \caption{A case illustrating a CLIP-T failure, underscoring a major issue in MagicBrush's evaluation approach. Despite VQGAN+CLIP generating a low-quality edit, it receives a higher CLIP score than the human-selected ground truth image, raising doubts on the evaluation's reliability. In contrast, our benchmark rates the ground truth edit significantly higher (with a score of 0.87) than the VQGAN-CLIP output (0.14). Additionally, preservation scores differ greatly - the ground truth nears perfect scores, while VQGAN-CLIP performs poorly except in subject size and position preservation.}
  \label{fig:magicbrush_clip}
\end{figure}

\begin{enumerate}
    \item \textbf{Absence of Ground Truth Images:} In MagicBrush, image quality is assessed using L1 and L2 distances between generated and ground-truth images, cosine similarity between CLIP and DINO \cite{dino}  embeddings, and CLIP-T \cite{dreambooth} for prompt-image alignment. However, this method presents two challenges. Firstly, it requires manual generation of ground-truth images using another editing model like DALL-E or tools like Photoshop, resulting in additional costs and impeding scalability. Secondly, MagicBrush's evaluation approach only considers their ground-truth images, overlooking other potentially correct edits. In contrast, our method assigns high scores to all potentially accurate edited images.
    
    \item \textbf{Unreliability of CLIP Scores:} The CLIP model's reliability has been called into question, as discussed in \cite{goel2022cyclip} and exemplified in Figure \ref{fig:magicbrush_clip}. 
    Furthermore, CLIP has demonstrated limited capabilities in evaluating certain edit instructions, especially those involving spatial alterations \cite{gokhale2023benchmarking}.
    
    \item \textbf{Detailed Edit-Specific Scores:} While MagicBrush offers comprehensive evaluation across various prompts with an edit-type independent pipeline, our benchmark specializes in object-oriented images, assessing only 9 edit types. However, it offers advantages such as exhaustive evaluation covering subject and background preservation, along with tailored logic for each edit type, ensuring nuanced and precise evaluation criteria beyond reliance solely on CLIP.
\end{enumerate}

\section{Conclusion}




PixLens offers a framework to evaluate different state-of-the-art models for image editing, designed to adapt seamlessly to any predefined evaluation dataset for instruction-based models. With a meticulous evaluation logic covering eight distinct edit types, PixLens employs detection and segmentation techniques to comprehensively assess the quality of the generated images. We employed our framework to evaluate six models over the EditVAL dataset. Notably, our findings reveal that no model excels across all edit types, with spatial manipulations posing a particular challenge for current image editing models. Demonstrating a superior correlation to human preference compared to EditVAL, PixLens presents a more accurate evaluation without depending on the CLIP metric, which has been proven unreliable in certain cases. Furthermore, by integrating disentanglement evaluation alongside traditional assessment metrics, PixLens offers a more comprehensive understanding of the performance of image editing models, setting a new standard for evaluation methodologies in this domain.

Our benchmark provides a crucial task-dependent evaluation, addressing the limitations of existing models that may overfit to the aforementioned metrics. This ensures a more accurate assessment of image-editing capabilities and aligns closely with the primary objectives of the task. PixLens aims to establish a new standard for evaluating text-guided image editing, marking a significant advancement in image quality assessment methodologies.


\subsection{Limitations}

In considering the limitations of PixLens, \emph{reliance on detection and segmentation models} presents a significant challenge, as these models can produce false positives and false negatives (as illustrated in the Supplementary Material)
impacting evaluation accuracy, particularly in cases like \texttt{object-removal}. Encoding subjective edit types such as \emph{closeness or proximity} is also problematic, as determining criteria for assessing these edits objectively is difficult. Additionally, capturing edits related to \emph{action} or \emph{viewpoint} changes is challenging for detection models, leading to evaluation inaccuracies. Evaluation of \emph{color edits} is biased towards pure basic colors, and despite efforts to mitigate this bias, assessing color edits remains subjective due to the lack of consensus in the field. Finally, while PixLens evaluates a substantial number of edit types (9 in total), there are still \emph{limitations in the variety of edits} it can assess, potentially restricting its applicability to certain image editing tasks and scenarios. Recognizing these limitations is crucial for understanding the scope and applicability of PixLens in evaluating image editing models. Ongoing efforts to address these limitations will enhance the benchmark's effectiveness and utility in the future.

\bibliographystyle{splncs04}
\bibliography{main}
\newpage

\usetikzlibrary{shapes.geometric, arrows}
\usetikzlibrary{shapes.multipart}
\usetikzlibrary{calc}
\usetikzlibrary{fit, positioning}

\definecolor{datasetGreen}{HTML}{93C47D}
\definecolor{baseBlue}{HTML}{CFE2F3}
\definecolor{componentOrange}{HTML}{F6B26B}

\definecolor{umlYellow}{HTML}{fffdcc}
\definecolor{umlPurple}{HTML}{ccccff}

\setlength{\fboxrule}{1pt}
\setlength{\fboxsep}{0pt}

\tikzset{font=\fontfamily{lmtt}\selectfont}
\tikzstyle{arrow} = [very thick, rounded corners, ->,>=stealth]

\tikzstyle{base} = [
    rectangle,
    rounded corners,
    minimum width=2cm, 
    minimum height=2cm,
    text centered, 
    draw=black, 
    very thick,
    fill=umlPurple
]

\tikzstyle{image} = [
    rectangle,
    draw=black,
    very thick,
    inner sep=0,
    outer sep=0
]







\section*{Supplementary Material}
This supplementary appendix provides additional detailed information on the evaluation methodologies utilized to assess various image editing models. The initial section discusses different edit types, offering specific algorithmic representations and in-depth discussions on the rationale behind each edit type evaluation method. Additionally, we present examples illustrating both successful and unsuccessful edits, along with the corresponding evaluation scores obtained using PixLens.

Furthermore, the appendix outlines the preprocessing steps applied to the MagicBrush dataset for PixLens evaluation and offers additional insights into subject preservation and disentanglement assessments. Lastly, we address common failure cases observed in detection models that may limit PixLens' evaluation capabilities, concluding this supplementary material.

\section{Evaluation Methods for Different Edit Types}

In this section, we offer a comprehensive description and pseudo-code for each of the edit types considered in our benchmark. These methods generate the edit-specific scores for the instances under evaluation.

\paragraph{Notation.} We adopt the same prompt format and notation as EditVAL \cite{basu2023editval}. Specifically, when referencing a specific image and editing operation, we denote the class of the image in the COCO dataset as \texttt{category}. Additionally, we use the \texttt{from} and \texttt{to} attributes from EditVAL. For instance, for an edit of type \texttt{object-replacement} over an image of class ``bench'', and prompt ``Replace the  woman with a kid'', the ``bench'' would be the \texttt{category}, and ``woman'' and ``kid'' would be the \texttt{from} and \texttt{to} attributes, respectively.

Furthermore, we refer to the input and edited images as $\mathbf{I_0}$ and $\mathbf{I_1}$, respectively $\mathbf{M}$ for
segmentation masks. The function \texttt{Detect} outputs a list of detected objects and the corresponding masks $\textbf{M}$. If a variable is not used in the algorithms, it is omitted.

\subsection{Object Addition}   

From what we have been able to observe in other attempts to benchmark text-guided image editing models, there is no established consensus regarding the interpretation of the instruction ``add an object'' in image manipulation operations. Some argue that the operation should be deemed successful only if the final image contains both the added object and the original category. Conversely, others contend that merely making the object visible in the image should suffice. After careful consideration, we have chosen not to impose any constraints on the detection of the image category, as subject preservation evaluation already verifies the retention of the main features of the primary image category. Consequently, to accommodate models that interpret the target as simply adding a new object in the image, we do not require the detection of the image category in the evaluation process when computing the edit-specific score.


\begin{algorithm}[]
   \caption{Object Addition}
   \label{object_addition}
\begin{algorithmic}
    \Require $\mathbf{I_1}$, \texttt{to}
    \State \texttt{to\_in\_edited} $\gets$ \texttt{Detect}($\mathbf{I_1}$, \texttt{to})
    \State \textbf{return} \texttt{score = exists(to\_in\_edited)}
    \end{algorithmic}
\end{algorithm}

\subsection{Size Change}

To evaluate size change edits, we start by detecting and segmenting the object targeted for resizing in both the input and edited images. Next, we validate whether the size change corresponds to the specified direction in the prompt. For instance, if the edit involves enlarging the object, we ensure that the area of the segmentation mask has increased in the edited image compared to the original. The magnitude of the size change does not affect the score, as our prompts do not dictate a specific new size. However, we enforce a minimum change $\delta$, set to $\delta = 0.1$ by default, relative to the original object. Additionally, we assess if the object's relative position remains relatively unchanged, factoring this into the final score. This involves checking if the smaller mask area (\eg~the input image's mask in enlargement) is contained within a larger area (\eg~the edited image's mask in enlargement) using a confidence threshold $\mathcal{T}$ (set to 0.9 by default). If the ratio of the intersection area to the smaller area exceeds the confidence threshold, the size change edit meets the expected criteria. While some may argue this aspect is covered in subject preservation evaluation, our experiments show this refinement enhances PixLens' alignment with human evaluation standards in size edits.

\begin{algorithm}[]
   \caption{Size Change}
   \label{alg:size_change}
\begin{algorithmic}
   \Require $\mathbf{I_0}, \mathbf{I_1}, \delta, \mathcal{T}$, \texttt{to}, \texttt{category}
   \State $\mathbf{M_0}$ $\gets$ \texttt{Detect}($\mathbf{I_0}$, \texttt{category})
   \State $\mathbf{M_1}$ $\gets$ \texttt{Detect}($\mathbf{I_1}$, \texttt{category})
   \If{\texttt{to == "small"} \textbf{and} \texttt{Area}($\mathbf{M_1}$) $/$ \texttt{Area}($\mathbf{M_0}$) $\geq 1 - \delta$}
   \State \textbf{return} \texttt{score} $=0$
   \ElsIf{\texttt{to == "big"} \textbf{and} \texttt{Area}($\mathbf{M_1}$) $/$ \texttt{Area}($\mathbf{M_0}$) $\leq 1 + \delta$}
   \State \textbf{return} \texttt{score} $=0$
   \EndIf
   
   \State $\mathbf{M_\texttt{small}} \gets \argmin_{\mathbf{M} \in \{\mathbf{M_0}, \mathbf{M_1}\}}$ \texttt{Area}$(\mathbf{M})$
   \If{$\texttt{Area}(\mathbf{M_0} \cap \mathbf{M_1}) \hspace{0.25em} / \hspace{0.25em} \texttt{Area}(\mathbf{M_\texttt{small}}) > \mathcal{T}$}
   \State \textbf{Return} \texttt{score} $= 1$
   \EndIf
   \State \textbf{Return} \texttt{score} $= 0$    
\end{algorithmic}
\end{algorithm}


\subsection{Positional Addition}

\begin{figure}
    \centering
    \includegraphics[width=0.8\linewidth]{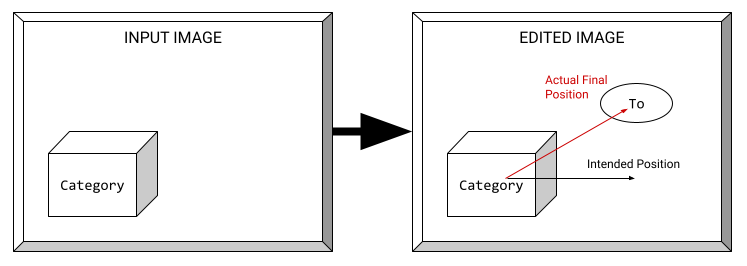} 
    \caption{\texttt{positional-addition} edit evaluation scheme, illustrating an edit where the intention is to add an object ``to the right of'' the \texttt{category} object.}
    \label{fig:positional-addition}
\end{figure}

This edit type involves adding a specific object (defined by the \texttt{to} attribute) to the image at a designated position relative to the \texttt{category} object (\cref{fig:positional-addition} illustrates the scheme for this edit type). Initially, our method identifies the intended relative position of the object by analyzing the editing prompt. For example, if the prompt is ``Add a bag on top of the table'', we determine the 2-dimensional vector ($\vec{v}$) that best describes the added bag's center of mass in relation to the table's center of mass (in this case, $\vec{v} = [\Delta_{x}, \Delta_{y}] = [0, 1]$). Next, we calculate the actual movement vector of the \texttt{category} object from its center of mass in the input image to the \texttt{to} object's center of mass in the edited image. If this computed vector has a negligible magnitude (indicating that the object was added directly at the category's position), the edit-specific score is set to zero. Otherwise, the score is determined based on the angle $\alpha$ between the computed movement vector and the unit vector representing the intended direction of movement described in the editing prompt. Specifically, the edit-specific score ranges from $1$ for $\alpha = 0\degree$ to $0$ for $\alpha = 90\degree$, with scores below $\alpha \geq 90\degree$ set to $0$. If the added object moves in the opposite direction of the intended position, the score is clipped to zero to avoid negative interpolation values.

\begin{algorithm}[]
\caption{Positional Addition}
\label{alg:positional_addition}
\begin{algorithmic}
\Require $\mathbf{I_0}, \mathbf{I_1}, \texttt{to}$
\State $\mathbf{M_0}$ $\gets$ \texttt{Detect}($\mathbf{I_0}$, \texttt{category})
\State $\mathbf{M_1}$ $\gets$ \texttt{Detect}($\mathbf{I_1}$, \texttt{to})
\State $\mathbf{cm_0}$ $\gets$ \texttt{center\_of\_mass}($\mathbf{M_0}$)
\State $\mathbf{cm_1}$ $\gets$ \texttt{center\_of\_mass}($\mathbf{M_1}$)
\State $\mathbf{dir}$ $\gets$ $\mathbf{cm_1} - \mathbf{cm_0}$
\State $\mathbf{ref}$ $\gets$ \texttt{get\_unit\_vector\_for\_direction}(\texttt{to})
\State $\alpha$ $\gets$ $\cos^{-1}\left(\frac{\mathbf{dir} \cdot \mathbf{ref}}{|\mathbf{dir}| |\mathbf{ref}|}\right)$
\State \textbf{return} $\texttt{score} = \max({0, \frac{90 - \alpha}{90}})$
\end{algorithmic}
\end{algorithm}

\subsection{Position Replacement}\label{sec:appendix_pos_replacement}

A successful \texttt{position-replacement} edit operation should alter the relative and absolute position of the \texttt{category} object within the image while preserving its main attributes (color, size, etc.), as well as the rest of the image (background preservation evaluation).

Evaluation of this edit type presents inherent complexity due to multiplicity. When the input image contains multiple instances of the category object, it's unclear if the editing model should adjust the position of all instances or only one. Additionally, determining which instance to adjust becomes challenging if only one change is expected. In cases where hallucinations occur, creating additional instances of the category object, deciding which instance to evaluate poses additional issues.

To automate assessment, we judge whether the position of the largest object (using \texttt{MultiplicityHandler.LARGEST}) in the input image has been correctly modified. Empirical testing showed this approach yields meaningful scores with current prompt-guided image editing models.

The evaluation procedure follows a similar process to \texttt{positional-addition} logic. However, we not only assess the relative position of the category object but also verify its final absolute position within the image. For example, if the prompt is ``Move the ball from the left to the right of the image'', we divide the image horizontally into three sections (left, center, and right) and confirm the final center of mass is in the correct section. 

\begin{algorithm}[]
\caption{Positional Replacement}
\label{alg:positional_replacement}
\begin{algorithmic}
\Require $\mathbf{I_0}, \mathbf{I_1}$, \texttt{init\_pos}, \texttt{intended\_rel\_pos}, \texttt{category}, \texttt{to}
\State $\mathbf{M_0}$ $\gets$ \texttt{Detect}($\mathbf{I_0}$, \texttt{category}).\texttt{largest}
\State \texttt{category\_in\_edited}$,\mathbf{M_1}$ $\gets$ \texttt{Detect}($\mathbf{I_1}$, \texttt{category}).\texttt{largest}
\If{!\texttt{category\_in\_edited} }
\State \textbf{Return} $\texttt{score} = 0$
\EndIf
\State $\mathbf{cm_0}$ $\gets$ \texttt{center\_of\_mass}($\mathbf{M_0}$)
\State $\mathbf{cm_1}$ $\gets$ \texttt{center\_of\_mass}($\mathbf{M_1}$)
\State $\mathbf{dir}$ $\gets$ $\mathbf{cm_1} - \mathbf{cm_0}$
\State $\texttt{relative\_score}\gets$ \texttt{compute\_relative}($\texttt{dir}$, $\texttt{init\_pos}$, $\texttt{intended\_rel\_pos}$)
\State $\texttt{absolute\_score}$ $\gets$ \texttt{compute\_absolute}($\texttt{img\_width}$, $\texttt{init\_pos}$, $\texttt{intended\_pos},\mathbf{cm_1})$
\If{$\texttt{relative\_score} > 0$}
\State \textbf{Return} $\texttt{score} = (\texttt{relative\_score} + \texttt{absolute\_score}) \hspace{0.25em} / \hspace{0.25em} 2$
\EndIf
\State \textbf{Return} $\texttt{score} = 0$
\end{algorithmic}
\end{algorithm}

\subsection{Object Replacement}

This type of edit aims to replace the \texttt{from} object with a new object \texttt{to}. The corresponding editing prompt structure is ``replace \texttt{from} with \texttt{to}''. Similar to other edit operations, evaluating this type poses challenges when multiple \texttt{from} and \texttt{to} objects are present in the original or edited image. To address this, we utilize the multiplicity handler. In this case, we select the input object with the largest area (\texttt{MultiplicityHandler.LARGEST}) and the edited object that is closest to the input one (\texttt{MultiplicityHandler.CLOSEST}). Followingly, we compute the intersection of the masks, returning a score of $1$ if non-empty, indicating success. Otherwise, we consider it a failed edit.

\begin{algorithm}[]
\caption{Object Replacement}
\label{alg:object_replacement}
\begin{algorithmic}
\Require $\mathbf{I_0}, \mathbf{I_1}$
\State $\mathbf{M_0}$ $\gets$ \texttt{Detect}($\mathbf{I_0}$, \texttt{category}).\texttt{largest}
\State $\mathbf{M_1}$ $\gets$ \texttt{Detect}($\mathbf{I_1}$, \texttt{to}).\texttt{closest}
\State $\mathbf{Intersection}$ $\gets$ $\mathbf{M_0} \cap \mathbf{M_1}$
\If{\texttt{area}($\mathbf{Intersection}$) $> 0$}
\State \textbf{Return} $\texttt{score} = 1$
\EndIf
\State \textbf{Return} $\texttt{score} = 0$
\end{algorithmic}
\end{algorithm}

\subsection{Object Removal}

The goal of this edit type is to evaluate how effectively a specified object category is removed from an image. We achieve this by comparing the number of detected \texttt{category} instances in both the input and edited images. This comparison allows us to determine the success of removing the specified object class. The edit-specific score is computed as the maximum of zero and one minus the ratio of the number of target categories in the edited image to the number in the input image, effectively quantifying the proportion of removed categories.

\begin{algorithm}
\caption{Object Removal}
\label{alg:object_removal}
\begin{algorithmic}
\Require $\mathbf{I_0}, \mathbf{I_1}, \texttt{category}$
\State $\texttt{category\_in\_input}$ $\gets$ \texttt{Detect}($\mathbf{I_0}$, \texttt{category})
\State $\texttt{category\_in\_edited}$ $\gets$ \texttt{Detect}($\mathbf{I_1}$, \texttt{category})
\State $\texttt{num\_input}$ $\gets$ \texttt{len}($\texttt{category\_in\_input}$)
\State $\texttt{num\_edited}$ $\gets$ \texttt{len}($\texttt{category\_in\_edited}$)
\State $\texttt{score}$ $\gets$ $\max\left(1 - \frac{\texttt{num\_edited}}{\texttt{num\_input}}, 0\right)$
\State \textbf{Return} $\texttt{score}$
\end{algorithmic}
\end{algorithm}

\subsection{Alter Parts}

The aim of this edit type is to modify certain parts of the \texttt{category} object in the image by adding complementary elements. For instance, an instruction could be: ``Add tomato toppings to the pizza'', where ``pizza'' represents the \texttt{category} and ``tomato toppings'' signifies the \texttt{to} attribute. Unlike the \texttt{object-addition} operation, the position of the \texttt{to} object is crucial, requiring intersection with the \texttt{category}. Furthermore, considering the possibility of multiple \texttt{category} objects in the input image, we ensure that all added ``complements'' intersect appropriately with each instance.

\begin{algorithm}[]
\caption{Alter Parts}
\label{alg:alter_parts}
\begin{algorithmic}
\Require $\mathbf{M_0}, \mathbf{M_1}$
\State $\texttt{cat\_objs}$ $\gets$ \texttt{Detect}($\mathbf{I_0}$, $\texttt{category}$)
\State $\texttt{to\_objs}$ $\gets$ \texttt{Detect}($\mathbf{I_1}$, $\texttt{to}$)
\State $\texttt{cm\_cat}$ $\gets$ \texttt{center\_of\_mass}($\mathbf{M_0}$, $\texttt{cat\_objs}$)
\State $\texttt{cm\_to}$ $\gets$ \texttt{center\_of\_mass}($\mathbf{M_1}$, $\texttt{to\_objs}$)
\State $\texttt{counter}$ $\gets$ $0$
\For{$\texttt{cat\_obj}$ in $\texttt{cat\_objs}$}
\State $\texttt{cm}$ $\gets$ $\texttt{cm\_cat}[\texttt{cat\_obj}]$
\State $\texttt{to\_obj}$ $\gets$ $\texttt{argmin}_{\texttt{obj} \in \texttt{to\_objs}} {||\texttt{cm\_to}[\texttt{obj}] - \texttt{cm}||_2}$
\State $\texttt{intersection}$ $\gets$ $\mathbf{M_1}[\texttt{to\_obj}]\cap{\mathbf{M_0}[\texttt{cat\_obj}]}$
\If {area$(\texttt{intersection}) > 0$}
    \State $\texttt{counter} \gets \texttt{counter} + 1$
\EndIf
\EndFor
\State \textbf{Return} \texttt{counter} $/$ \texttt{len}($\texttt{cat\_objs}$)
\end{algorithmic}
\end{algorithm}

\subsection{Color}\label{sec:appendix_color}

The objective here is to change the color of the main object in the image while ideally preserving the color of the background and other elements (which is thoroughly evaluated in the background preservation assessment). To achieve this, we generate an image with pixels of the target color using the \texttt{ImageColor} module from the \texttt{Pillow} python library\footnote{Note: This approach limits the variety of colors for evaluation in PixLens.}. We then compute 1D histograms for the masked edited image and the target color image over each RGB channel. These histograms are smoothed using a Gaussian kernel to reduce noise. A correlation analysis is then performed between the smoothed target color histograms and the histograms of the masked image. Finally, the mean correlation score across all three RGB channels is computed and outputted as the edit-specific score.

\begin{algorithm}[]
\caption{Color}
\label{alg:color}
\begin{algorithmic}
\State $\texttt{masked\_image} \gets$ \texttt{masked edited image}
\State $\texttt{target\_color\_img} \gets$ \texttt{image of same size with target color pixels}
\State $\texttt{r}_\texttt{target}, \texttt{g}_\texttt{target}, \texttt{b}_\texttt{target} \gets$ \texttt{Generate 1D-histograms of} $\texttt{target\_color\_img}$
\State  $\texttt{r}_\texttt{target}, \texttt{g}_\texttt{target}, \texttt{b}_\texttt{target} \gets$ \texttt{GaussianKernel}( $\texttt{r}_\texttt{target}, \texttt{g}_\texttt{target}, \texttt{b}_\texttt{target}$)
\State $\texttt{corr}_r, \texttt{corr}_g, \texttt{corr}_b \gets$ \texttt{Correlation scores}
\State $\texttt{score} \gets$ \texttt{Mean}($\texttt{corr}_r, \texttt{corr}_g, \texttt{corr}_b$)
\State \textbf{Return} $\texttt{score}$
\end{algorithmic}
\end{algorithm}

\section{PixLens Visual Demonstrations}

\begin{figure}
  \centering
  \begin{subfigure}{0.23\textwidth}
    \centering
    \includegraphics[width=\linewidth]{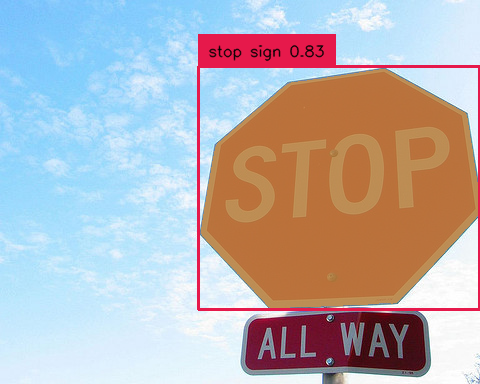}
    \caption{Original Image A}
  \end{subfigure}
  \begin{subfigure}{0.23\textwidth}
    \centering
    \includegraphics[width=\linewidth]{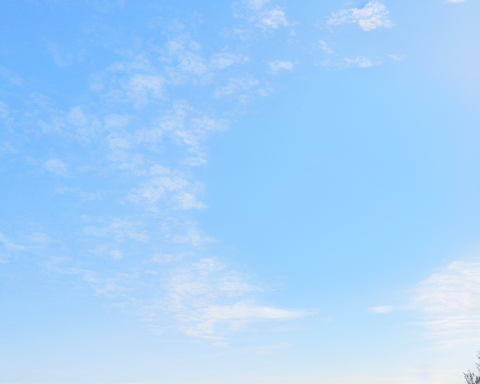}
    \caption{Edited Image A}
  \end{subfigure}
  \hspace{0.20cm}
  \begin{subfigure}{0.23\textwidth}
    \centering
    \includegraphics[width=\linewidth]{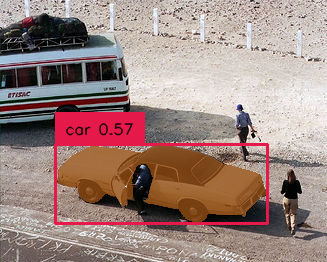}
    \caption{Original Image B}
  \end{subfigure}
  \begin{subfigure}{0.23\textwidth}
    \centering
    \includegraphics[width=\linewidth]{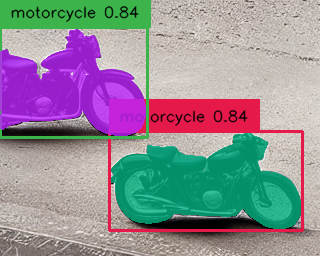}
    \caption{Edited Image B}
  \end{subfigure}
  
  \vspace{0.25cm}
  
  \centering
  \begin{subfigure}{0.23\textwidth}
    \centering
    \includegraphics[width=\linewidth]{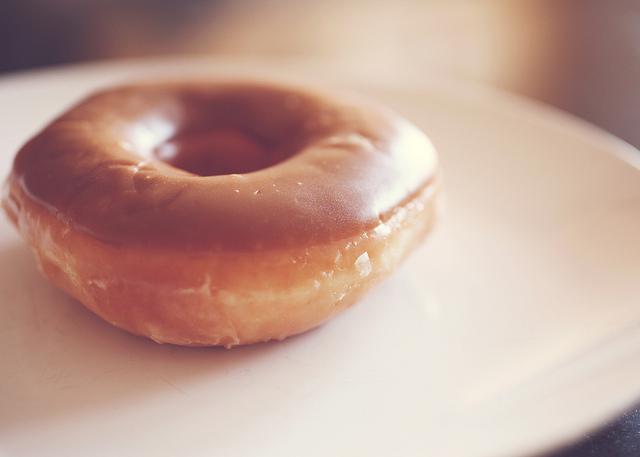}
    \caption{Original Image C}
  \end{subfigure}
  \begin{subfigure}{0.23\textwidth}
    \centering
    \includegraphics[width=\linewidth]{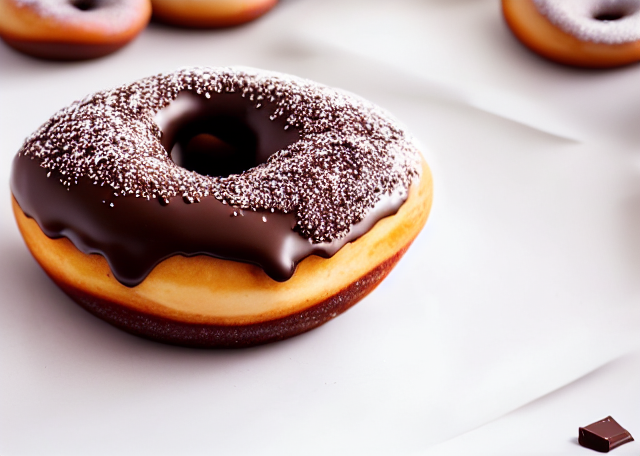}
    \caption{Edited Image C}
  \end{subfigure}
  \hspace{0.20cm}
  \begin{subfigure}{0.23\textwidth}
    \centering
    \includegraphics[width=\linewidth]{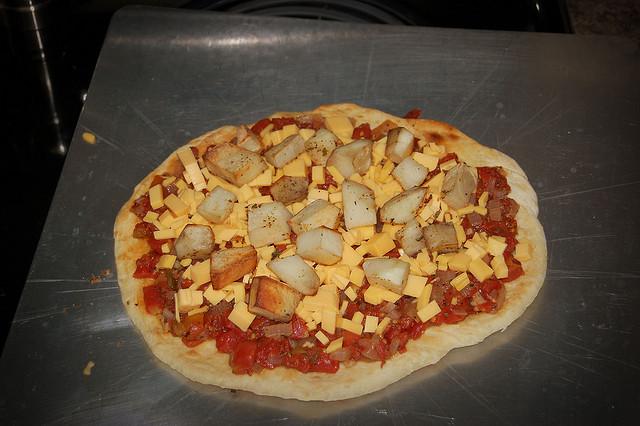}
    \caption{Original Image D}
  \end{subfigure}
  \begin{subfigure}{0.23\textwidth}
    \centering
    \includegraphics[width=\linewidth]{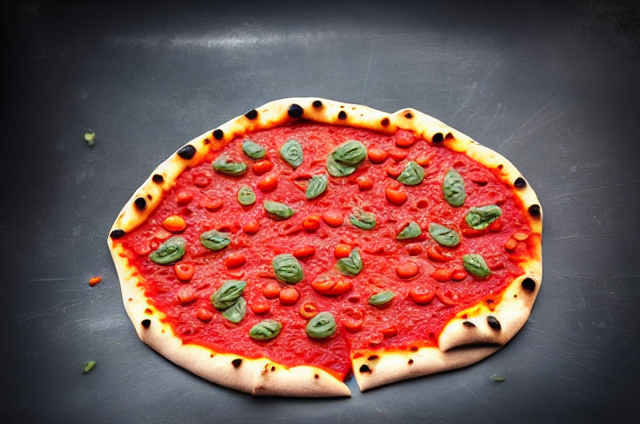}
    \caption{Edited Image D}
  \end{subfigure}
  \caption{A: ``Remove the stop sign'', B: ``Replace the car with a motorbike'', C: ``Add chocolate toppings to the donut'', D: ``Add tomato toppings to the pizza''. All the edits performed with InstructPix2Pix.}
  \label{fig:successful_edits}
\end{figure}

We showcase the effectiveness of our evaluation benchmark in assessing various image editing models by examining both successful and unsuccessful edits to understand how PixLens scores and artifacts accurately reflect editing quality. While we primarily focus on InstructPix2Pix due to its high performance, we present four instances of successful edits in \cref{fig:successful_edits} (all four attain a perfect edit-specific score of 1).

In Case A, an \texttt{object-removal} edit is performed to eliminate a stop sign, resulting in the successful removal of the sign with minimal hallucinations. This edit achieves a high background preservation score of 0.77. In Case B, an \texttt{object-replacement} edit replaces a car with a motorcycle, although some minor inaccuracies occur in the replacement, leading to a slightly lower background preservation score of 0.73. Cases C and D correspond to \texttt{alter-parts} edits, where chocolate toppings are added to a donut and tomato toppings to a pizza, respectively. Both edits achieve high scores, with background preservation scores of 0.76 and 0.81, respectively.

\begin{figure}
  \centering
  \begin{subfigure}{0.23\textwidth}
    \centering
    \includegraphics[width=\linewidth]{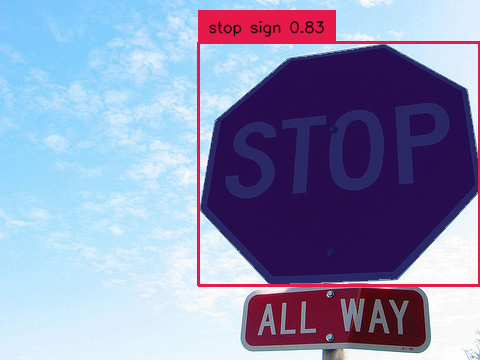}
    \caption{Original Image A}
  \end{subfigure}
  \begin{subfigure}{0.23\textwidth}
    \centering
    \includegraphics[width=\linewidth]{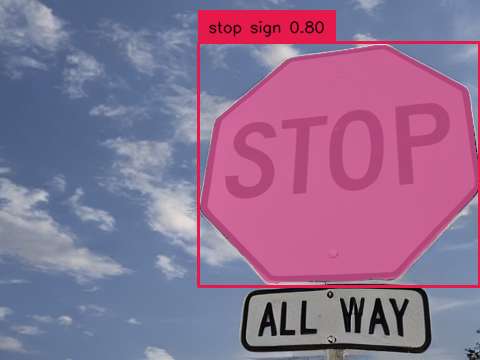}
    \caption{Edited Image A}
  \end{subfigure}
  \hspace{0.20cm}
  \begin{subfigure}{0.23\textwidth}
    \centering
    \includegraphics[width=\linewidth]{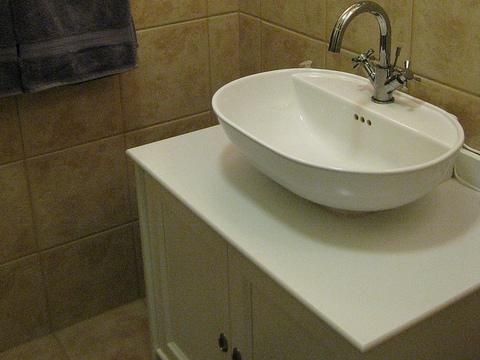}
    \caption{Original Image B}
  \end{subfigure}
  \begin{subfigure}{0.23\textwidth}
    \centering
    \includegraphics[width=\linewidth]{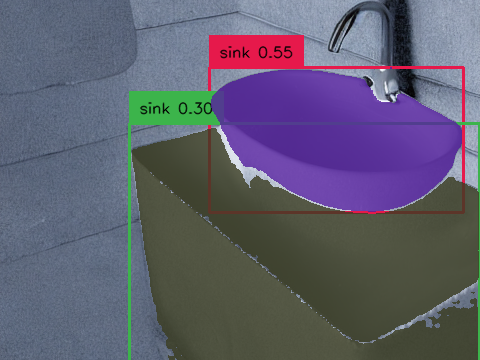}
    \caption{Edited Image B}
  \end{subfigure}
  
  \vspace{0.25cm}
  
  \centering
  \begin{subfigure}{0.23\textwidth}
    \centering
    \includegraphics[width=\linewidth]{Pixlens_ECCV_Appendix_Workshop/figs/failures/failure3.jpg}
    \caption{Original Image C}
  \end{subfigure}
  \begin{subfigure}{0.23\textwidth}
    \centering
    \includegraphics[width=\linewidth]{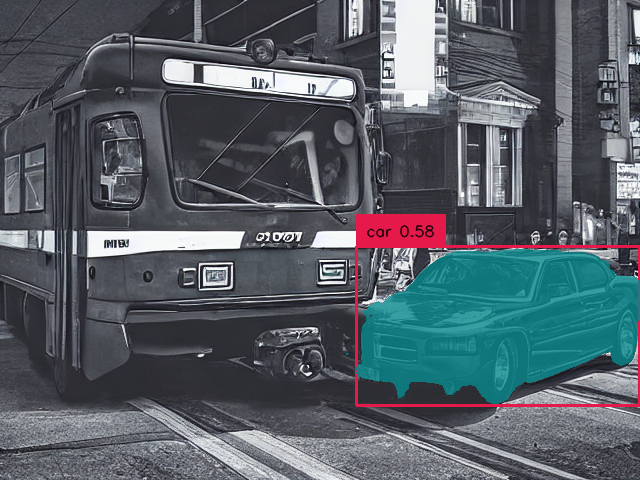}
    \caption{Edited Image C}
  \end{subfigure}
  \hspace{0.20cm}
  \begin{subfigure}{0.23\textwidth}
    \centering
    \includegraphics[width=\linewidth]{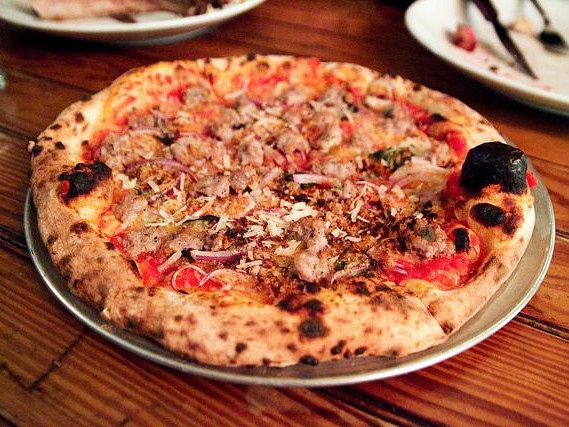}
    \caption{Original Image D}
  \end{subfigure}
  \begin{subfigure}{0.23\textwidth}
    \centering
    \includegraphics[width=\linewidth]{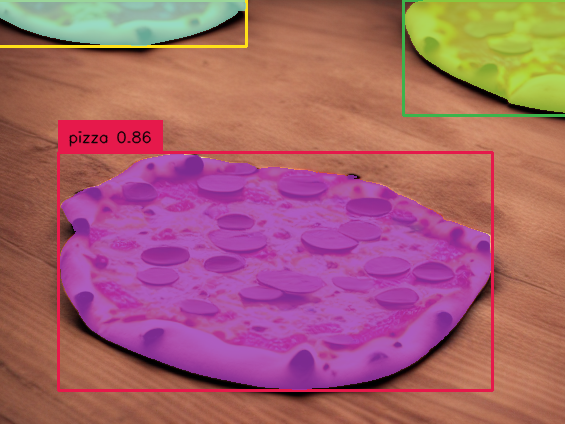}
    \caption{Edited Image D}
  \end{subfigure}
  \caption{A: ``Remove the stop sign'', B: ``Add a toothbrush'', C: ``Replace the car with a motorcycle'', D: ``Remove the pizza''. Images A, B and C are edited with ControlNet and Image C with InstructPix2Pix.}
  \label{fig:unsuccessful_edits}
\end{figure}

Now, let's delve into failed edits, illustrated in \cref{fig:unsuccessful_edits}. In Image A, an instance of \texttt{object-removal}, the stop sign is not correctly removed in the edited image, resulting in an edit-specific score of 0. Image B represents a failed \texttt{object-addition} edit, where the toothbrush is not correctly added to the image, also attaining an edit-specific score of 0. For Case C, the car is not replaced by a motorcycle, and in Case D, the pizza remains unchanged in the image, leading to both edits receiving a score of 0.

\newpage
\section{Preprocessing the MagicBrush Dataset for PixLens Evaluation}

In alignment with the methodology outlined in the main paper, we focused our preprocessing efforts on the ``dev'' subset of the MagicBrush dataset, comprising 528 edits alongside their corresponding input and ground truth images, as well as the respective prompts. To adapt these edits to our evaluation benchmark, we manually extracted and defined the necessary attributes for each edit, namely the \texttt{category}, \texttt{from}, and \texttt{to} attributes, based on the provided prompts.

\begin{figure}
    \begin{minipage}{\textwidth}
    \centering
    \begin{subfigure}[b]{0.35\textwidth} 
        \centering
        \includegraphics[scale=0.2]{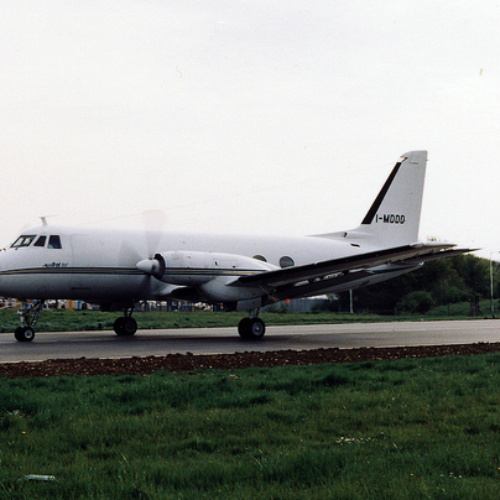}
    \end{subfigure}
    \begin{subfigure}[b]{0.35\textwidth} 
        \centering
        \includegraphics[scale=0.2]{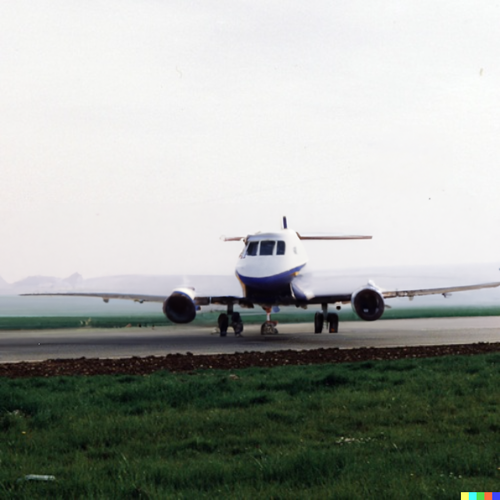}
    \end{subfigure}
    \centering
    \text{``Can we have a blue airplane?''}
    \end{minipage}
    \\ 
    \\
     \begin{minipage}{\textwidth}
     \centering
    \begin{subfigure}[b]{0.35\textwidth} 
        \centering
        \includegraphics[scale=0.2]{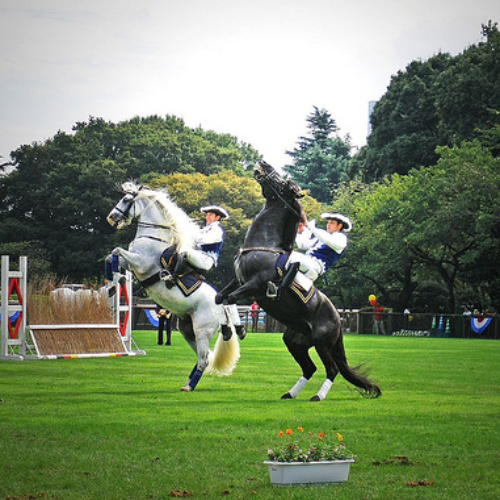}
    \end{subfigure}
    \centering
    \begin{subfigure}[b]{0.35\textwidth} 
        \centering
        \includegraphics[scale=0.2]{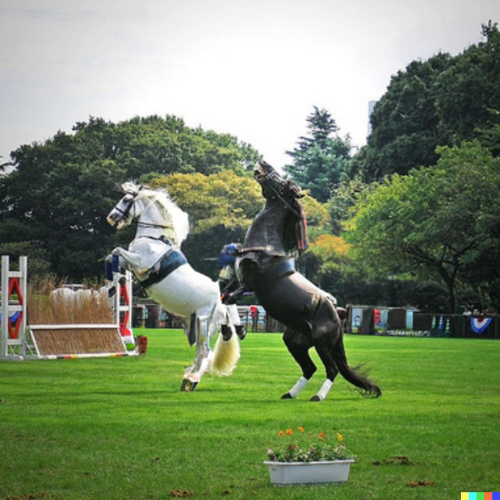}
    \end{subfigure}
    \\
    \text{``Make the two men fall''}
    \end{minipage}
    \caption{Examples of incorrect ground-truth instances in the MagicBrush ``dev'' subset. In the first instance, the edited image fails to depict a blue airplane as requested. In the second instance, the two men vanish instead of appearing to fall.}
    \label{fig:bad_gt}
\end{figure}

Additionally, we filtered out edit operations that did not meet our criteria. Specifically, we discarded edits in which: (i) the ground truth images inaccurately represented the intended edits, as demonstrated in \cref{fig:bad_gt}; and (ii) the edits could not be classified into any of the predefined edit types in our evaluation framework, such as edits involving spatial relationships (\eg~``Let the cow be closer to the farm'') or alterations related to action, viewpoint, or texture.

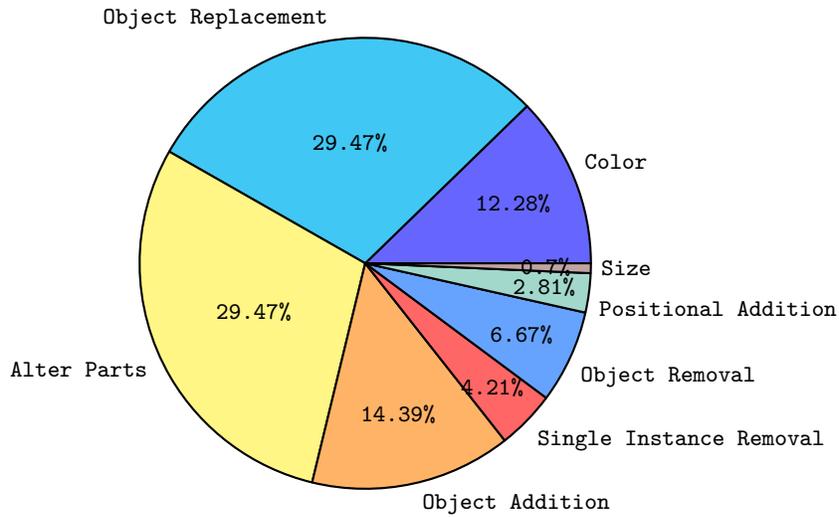
\begin{figure}
  \centering
  \begin{tikzpicture}
    \pie{12.28/Color, 29.47/Object Replacement, 29.47/Alter Parts, 14.39/Object Addition, 4.21/Single Instance Removal, 6.67/Object Removal, 2.81/Positional Addition, 0.7/Size}
  \end{tikzpicture}
  \caption{Edit type distribution in MagicBrush filtered subset}
  \label{fig:magicbrush-edit-distribution}
\end{figure}

In Table \ref{tab:magicbrush-edit-distribution} and Figure \ref{fig:magicbrush-edit-distribution}, we present the final distribution of MagicBrush editing operations over which the evaluation was run using PixLens. After filtering out unimplemented edit types and unreliable ground truths, we retained a total of 285 edit operations, with 260 of them successfully evaluated.

From the distribution, it's evident that \texttt{object-replacement} and \texttt{alter-parts} are the most prevalent edit types, followed by \texttt{object-addition} and \texttt{color} edits. Notably, there is an imbalance in the dataset, particularly with spatial manipulation-related edit types. For instance, \texttt{positional-addition} edits only comprise 8 instances, while \texttt{size} edits are represented by merely 2 examples. Furthermore, \texttt{position-replacement} edits are absent entirely from the dataset.


\begin{table}
    \caption{Post-filtering Edit Type Distribution in MagicBrush}
    \label{tab:magicbrush-edit-distribution}
    \vskip 0.1in
    \begin{center}
    \begin{scriptsize} 
    \begin{sc}
    \begin{tabular}{l@{\hspace{0.75em}}c@{\hspace{0.75em}}c}
        \toprule
        \textbf{Edit Type} & \textbf{Nº Edits} & \textbf{Nº Successfully Evaluated}  \\ \hline
        Color & 35 & 28  \\ 
        Object Replacement & 84 & 79  \\ 
        Alter Parts & 84 & 76  \\ 
        Object Addition & 41 & 40  \\ 
        Single Instance Removal & 12 & 12  \\ 
        Object Removal & 19 & 17  \\ 
        Positional Addition & 8 & 6  \\ 
        Size & 2 & 2  \\
        \hline
        Total & 285 & 260 \\
        \bottomrule
    \end{tabular}
    \end{sc}
    \end{scriptsize}
    \end{center}
    \vskip -0.1in
\end{table}

\section{Supplementary Information on Subject Preservation Evaluation}
Subject preservation evaluation entails computing five scores to assess the degree to which the subject is retained:

\begin{enumerate}
    \item \textbf{SIFT Score} $\uparrow$: This score represents the percentage of SIFT descriptors in the images that form good matches, determined using Lowe's ratio test \cite{SIFT}. Typically, values below 0.075 are considered poor, while values above 0.1 are indicative of good preservation. The SIFT score primarily tracks structural preservation.
    
    \item \textbf{Aligned IoU} $\uparrow$: The Intersection over Union (IoU) of the aligned masks, measuring changes in the size and shape of the subject. The masks are aligned by their first row and column (i.e. so that their top-left corners coincide).

    \item \textbf{SSIM} \cite{SSIM} $\uparrow$: The Structural Similarity Index Measure is a metric used to quantify the similarity between two images, focusing on luminance, contrast, and structure. It produces values ranging from -1 to 1, with 1 indicating perfect similarity.
    
    \item \textbf{Color Similarity} $\uparrow$: This score represents the mean of channel-wise histogram correlations, where values close to 1 indicate high color similarity.
    
    \item \textbf{Position Score} $\downarrow$: The position score reflects the distance between the centroids of the masks, normalized by the image size. Lower values, closer to 0, signify better preservation.
\end{enumerate}

It's important to interpret the subject preservation scores in the context of the specific edit type. For instance, the color similarity score may not be relevant when the edit prompt involves a directive like \texttt{``Make the bag red''}.

\section{Failure Cases of Detection Models}

In our analysis, we discovered that misalignments between the computed evaluation scores from our benchmark and human evaluation usually stem from the inaccuracies of the detection model \cite{groundedsam}, or from hallucination artifacts generated by the image editing models \cite{instructpix2pix, lcm, controlnet}. We categorize these instances into true negatives and false positives. We believe these issues will be significantly reduced with the development of improved zero-shot detection models, though addressing this is beyond the scope of our work.

\begin{figure}
    \centering
    \begin{subfigure}[t]{0.31\textwidth}
    \centering
    \includegraphics[width=\linewidth]{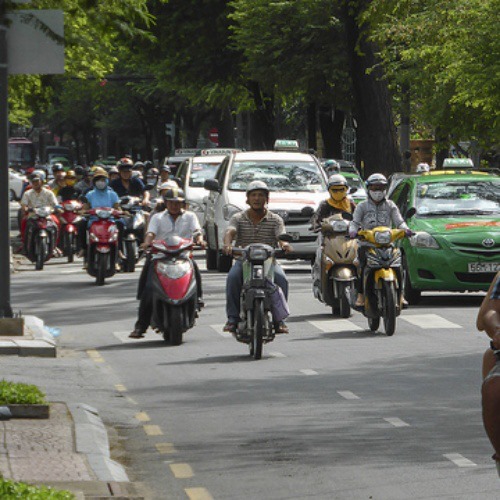}
    \caption{Original Image}
    \end{subfigure}
    \hfill
    \begin{subfigure}[t]{0.31\textwidth}
    \centering
    \includegraphics[width=\linewidth]{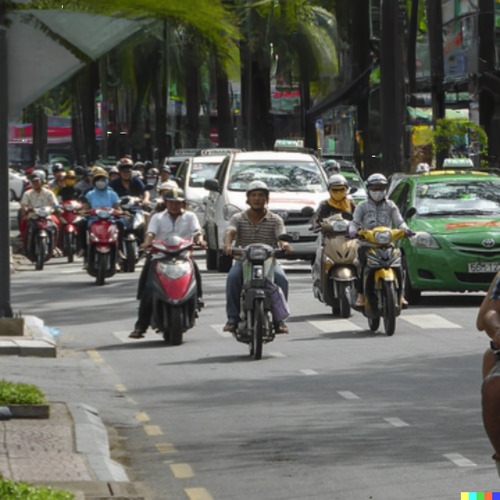}
    \caption{Edited Image}
    \end{subfigure}
    \hfill
    \begin{subfigure}[t]{0.31\textwidth}
    \centering
    \includegraphics[width=\linewidth]{Pixlens_ECCV_Appendix_Workshop/figs/palm_edited.jpeg}
    \caption{Annotated Image (No Detection)}
    \end{subfigure}
    \caption{Example of a \textbf{False Negative} case where the object detection model fails to detect the added object. (Specific Instruction: ``Replace the trees with palm trees'', Image Editing Model: ControlNet. Object Detection Model: Grounded-SAM)}
    \label{fig:lim1}
\end{figure}

\textbf{False Negatives:} In certain scenarios, PixLens may inaccurately assign low scores to well-executed edits, especially in the following scenarios:

\begin{enumerate}
    \item \emph{Object Replacement}: Although the image editing model correctly replaces the intended object, the detection model fails to identify it (see \cref{fig:lim1}).
    \item \emph{Color Change}: When the image editing model slightly alters the color of a multi-colored object, PixLens may assign a low score due to the presence of non-target color pixels, impacting the correlation analysis.
\end{enumerate}

\begin{figure}
    \centering
    \begin{subfigure}{0.31\textwidth}
    \centering
    \includegraphics[width=\linewidth]{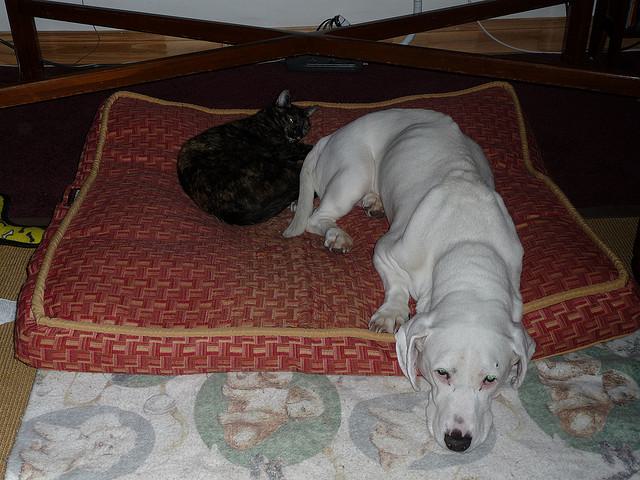}
    \caption{Original Image}
    \end{subfigure}
    \hfill
    \begin{subfigure}{0.31\textwidth}
    \centering
    \includegraphics[width=\linewidth]{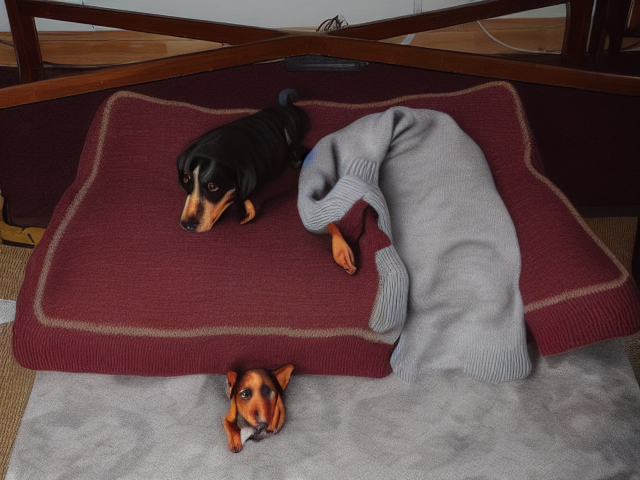}
    \caption{Edited Image}
    \end{subfigure}
    \hfill
    \begin{subfigure}{0.31\textwidth}
    \centering
    \includegraphics[width=\linewidth]{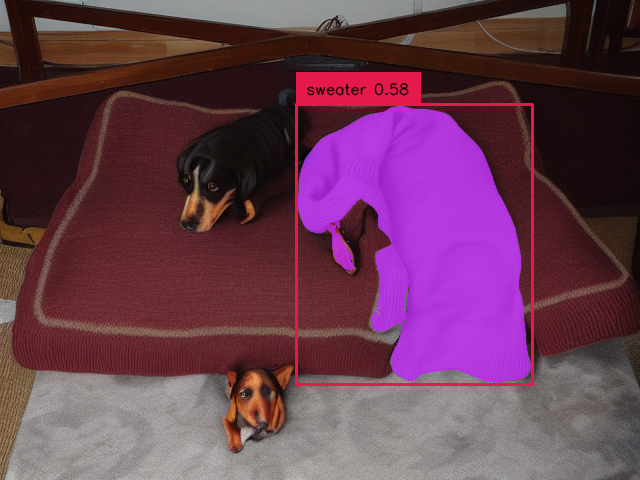}
    \caption{Annotated Image}
    \end{subfigure}
    \caption{Example of a \textbf{False Positive} case where the object detection model falsely detects the added artifact as a sweater. (Specific Instruction: ``Let the dog wear a sweater''. Image Editing Model: InstructPix2Pix. Object Detection Model: Grounded-SAM)}
    \label{fig:lim2}
\end{figure}

\textbf{False Positives:} Conversely, PixLens may assign high scores to poorly executed edits more prominently in the following cases:

\begin{enumerate}
    \item Edit types related to the \emph{addition} of an object: If the image editing model adds an object with correct texture but incorrect shape, the detection model might erroneously detect it due to its texture and color, even though a human observer would not recognize it as such (see \cref{fig:lim2}).
    \item \emph{Object Replacement}: When the image editing model replaces the original object but also introduces artifacts, the detection model might detect these artifacts as additional instances of the replacement object.
\end{enumerate}

\section{Supplementary Information on Disentanglement Evaluation}

\subsection{Generated Dataset}

In the generated dataset, latent representations are formed by combining attributes and objects. We have chosen four distinct attribute types for this benchmark: texture, color, style, and pattern. The available objects and attributes are as follows:

\begin{enumerate}
    \item \textbf{Objects}: Chair, Bowl, Plate, Nail, Bucket, Backpack, Book, Ball, Clock, Donut.
    \item \textbf{Textures}: Steel, Wood, Glass, Plastic, Wool, Cotton, Silk.
    \item \textbf{Colors}: Verdant, Red, Azure, Green, Gold, Purple, Black, Pink.
    \item \textbf{Styles}: Vintage, Modern, Abstract, Realistic, Cartoon, Surreal, Expressionist, Futuristic, Retro.
    \item \textbf{Patterns}: Striped, Polka Dot, Plaid, Paisley, Floral, Geometric, Abstract, Animal Print, Checkered, Herringbone.
\end{enumerate}

Therefore, examples of prompts for latent representations include ``Vintage chair'', ``Striped nail'', or ``Red plate''.

\subsection{Plots}

For the inter-attribute section of the disentanglement evaluation, following the dataset construction outlined in the main part, a linear classifier is trained for 50 epochs. To address computational constraints, we randomly sample seeded data for each class in the final dataset. The test size used to extract accuracy is set to 30\% of the total number of samples.

In addition to accuracy and balanced accuracy metrics, we provide confusion matrices on the test set for each model to offer deeper insights into attribute classification (refer to \cref{fig:first,fig:second,fig:third,fig:fourth}).

\begin{figure}
    \centering
    \begin{minipage}[b]{0.48\textwidth}
        \usetikzlibrary{matrix,calc}


\def\myConfMat{{
{105,    5,      0,      0},    
{   0,  136,    2,      0},    
{   0,   17,   87,      0},    
{   0,    3,      0,   91}     
}}

\def\classNames{{"Color","Pattern","Style","Texture"}} 

\def\numClasses{4} 

\def\myScale{1.1} 
\begin{tikzpicture}[
    scale = \myScale,
    ]

\tikzset{vertical label/.style={rotate=90,anchor=east}}   
\tikzset{diagonal label/.style={rotate=45,anchor=north east}}

\foreach \y in {1,...,\numClasses} 
{
    \node [anchor=east] at (0.4,-\y) {\pgfmathparse{\classNames[\y-1]}\pgfmathresult}; 
    
    \foreach \x in {1,...,\numClasses}  
    {
        \def\totSamples{0}
            \foreach \k in {1,...,\numClasses}
            {
                \pgfmathparse{\myConfMat[\y-1][\k-1]} 
                \xdef\totSamples{\totSamples+\pgfmathresult}
            }
            \pgfmathparse{\totSamples} \xdef\totSamples{\pgfmathresult} 
        
        \begin{scope}[shift={(\x,-\y)}]
            \def\mVal{\myConfMat[\y-1][\x-1]} 
            \pgfmathtruncatemacro{\r}{\mVal}   %
            \pgfmathtruncatemacro{\p}{round(\r/\totSamples*100)}
            \coordinate (C) at (0,0);
            \ifthenelse{\p<50}{\def\txtcol{black}}{\def\txtcol{white}} 
            \node[
                draw,                 
                text=\txtcol,         
                align=center,         
                fill=black!\p,        
                minimum size=\myScale*10mm,    
                inner sep=0,          
            ] (C) {\r\\\p\%};     
            \ifthenelse{\y=\numClasses}{
                \node [] at ($(C)-(0,0.75)$) 
                {\pgfmathparse{\classNames[\x-1]}\pgfmathresult};}{}
        \end{scope}
    }
}
\coordinate (yaxis) at (-0.9,0.5-\numClasses/2);  
\coordinate (xaxis) at (0.5+\numClasses/2, -\numClasses-1.25); 
\node [vertical label] at (yaxis) {Actual Class};
\node []               at (xaxis) {Predicted Class};
\end{tikzpicture}

        \caption{ControlNet}
        \label{fig:first}
    \end{minipage}
    \hfill
    \begin{minipage}[b]{0.48\textwidth}
        \usetikzlibrary{matrix,calc}


\def\myConfMat{{
{105,    5,      0,      0},    
{   0,  138,    0,      0},    
{   2,   20,   65,      17},    
{   0,    3,      0,   91}     
}}

\def\classNames{{"Color","Pattern","Style","Texture"}} 

\def\numClasses{4} 

\def\myScale{1.1} 
\begin{tikzpicture}[
    scale = \myScale,
    ]

\tikzset{vertical label/.style={rotate=90,anchor=east}}   
\tikzset{diagonal label/.style={rotate=45,anchor=north east}}

\foreach \y in {1,...,\numClasses} 
{
    \node [anchor=east] at (0.4,-\y) {\pgfmathparse{\classNames[\y-1]}\pgfmathresult}; 
    
    \foreach \x in {1,...,\numClasses}  
    {
        \def\totSamples{0}

        \foreach \k in {1,...,\numClasses}
        {
            \pgfmathparse{\myConfMat[\y-1][\k-1]} 
            \xdef\totSamples{\totSamples+\pgfmathresult}
        }
    \pgfmathparse{\totSamples} \xdef\totSamples{\pgfmathresult} 
        
        \begin{scope}[shift={(\x,-\y)}]
            \def\mVal{\myConfMat[\y-1][\x-1]} 
            \pgfmathtruncatemacro{\r}{\mVal}   %
            \pgfmathtruncatemacro{\p}{round(\r/\totSamples*100)}
            \coordinate (C) at (0,0);
            \ifthenelse{\p<50}{\def\txtcol{black}}{\def\txtcol{white}} 
            \node[
                draw,                 
                text=\txtcol,         
                align=center,         
                fill=black!\p,        
                minimum size=\myScale*10mm,    
                inner sep=0,          
            ] (C) {\r\\\p\%};     
            \ifthenelse{\y=\numClasses}{
                \node [] at ($(C)-(0,0.75)$) 
                {\pgfmathparse{\classNames[\x-1]}\pgfmathresult};}{}
        \end{scope}
    }
}
\coordinate (yaxis) at (-0.9,0.5-\numClasses/2);  
\coordinate (xaxis) at (0.5+\numClasses/2, -\numClasses-1.25); 
\node []               at (xaxis) {Predicted Class};
\end{tikzpicture}

        \caption{LCM}
        \label{fig:second}
    \end{minipage}

    \centering
    \begin{minipage}[b]{0.48\textwidth}
        \usetikzlibrary{matrix,calc}


\def\myConfMat{{
{105,    5,    0,    0},  
{0,    138,  0,    0},  
{2,      20,  65,    17},  
{0,      3,     0,  91},  
}}

\def\classNames{{"Color","Pattern","Style","Texture"}} 

\def\numClasses{4} 

\def\myScale{1.1} 
\begin{tikzpicture}[
    scale = \myScale,
    ]

\tikzset{vertical label/.style={rotate=90,anchor=east}}   
\tikzset{diagonal label/.style={rotate=45,anchor=north east}}

\foreach \y in {1,...,\numClasses} 
{
    \node [anchor=east] at (0.4,-\y) {\pgfmathparse{\classNames[\y-1]}\pgfmathresult}; 
    
    \foreach \x in {1,...,\numClasses}  
    {
        \def\totSamples{0}
            \foreach \k in {1,...,\numClasses}
            {
                \pgfmathparse{\myConfMat[\y-1][\k-1]} 
                \xdef\totSamples{\totSamples+\pgfmathresult}
            }
            \pgfmathparse{\totSamples} \xdef\totSamples{\pgfmathresult} 
        
        \begin{scope}[shift={(\x,-\y)}]
            \def\mVal{\myConfMat[\y-1][\x-1]} 
            \pgfmathtruncatemacro{\r}{\mVal}   %
            \pgfmathtruncatemacro{\p}{round(\r/\totSamples*100)}
            \coordinate (C) at (0,0);
            \ifthenelse{\p<50}{\def\txtcol{black}}{\def\txtcol{white}} 
            \node[
                draw,                 
                text=\txtcol,         
                align=center,         
                fill=black!\p,        
                minimum size=\myScale*10mm,    
                inner sep=0,          
            ] (C) {\r\\\p\%};     
            \ifthenelse{\y=\numClasses}{
                \node [] at ($(C)-(0,0.75)$) 
                {\pgfmathparse{\classNames[\x-1]}\pgfmathresult};}{}
        \end{scope}
    }
}
\coordinate (yaxis) at (-0.9,0.5-\numClasses/2);  
\coordinate (xaxis) at (0.5+\numClasses/2, -\numClasses-1.25); 
\node [vertical label] at (yaxis) {Actual Class};
\node []               at (xaxis) {Predicted Class};
\end{tikzpicture}

        \caption{InstructPix2Pix}
        \label{fig:third}
    \end{minipage}
    \hfill
    \begin{minipage}[b]{0.48\textwidth}
        \usetikzlibrary{matrix,calc}


\def\myConfMat{{
{105,    0,    0,    5},  
{9,    69,  40,    20},  
{0,      0,  87,    17},  
{0,      0,     0,  94},  
}}

\def\classNames{{"Color","Pattern","Style","Texture"}} 

\def\numClasses{4} 

\def\myScale{1.1} 
\begin{tikzpicture}[
    scale = \myScale,
    ]

\tikzset{vertical label/.style={rotate=90,anchor=east}}   
\tikzset{diagonal label/.style={rotate=45,anchor=north east}}

\foreach \y in {1,...,\numClasses} 
{
    \node [anchor=east] at (0.4,-\y) {\pgfmathparse{\classNames[\y-1]}\pgfmathresult}; 
    
    \foreach \x in {1,...,\numClasses}  
    {
        \def\totSamples{0}
        \foreach \k in {1,...,\numClasses}
        {
            \pgfmathparse{\myConfMat[\y-1][\k-1]} 
            \xdef\totSamples{\totSamples+\pgfmathresult}
        }
        \pgfmathparse{\totSamples} \xdef\totSamples{\pgfmathresult} 
        
        \begin{scope}[shift={(\x,-\y)}]
            \def\mVal{\myConfMat[\y-1][\x-1]} 
            \pgfmathtruncatemacro{\r}{\mVal}   %
            \pgfmathtruncatemacro{\p}{round(\r/\totSamples*100)}
            \coordinate (C) at (0,0);
            \ifthenelse{\p<50}{\def\txtcol{black}}{\def\txtcol{white}} 
            \node[
                draw,                 
                text=\txtcol,         
                align=center,         
                fill=black!\p,        
                minimum size=\myScale*10mm,    
                inner sep=0,          
            ] (C) {\r\\\p\%};     
            \ifthenelse{\y=\numClasses}{
                \node [] at ($(C)-(0,0.75)$) 
                {\pgfmathparse{\classNames[\x-1]}\pgfmathresult};}{}
        \end{scope}
    }
}
\coordinate (yaxis) at (-0.9,0.5-\numClasses/2);  
\coordinate (xaxis) at (0.5+\numClasses/2, -\numClasses-1.25); 
\node []               at (xaxis) {Predicted Class};
\end{tikzpicture}

        \caption{VQGAN+CLIP}
        \label{fig:fourth}
    \end{minipage}
\end{figure}

\subsection{Disentanglement intra-sample evaluation examples}
\begin{figure}
    \centering
    \includegraphics[scale = 0.3]{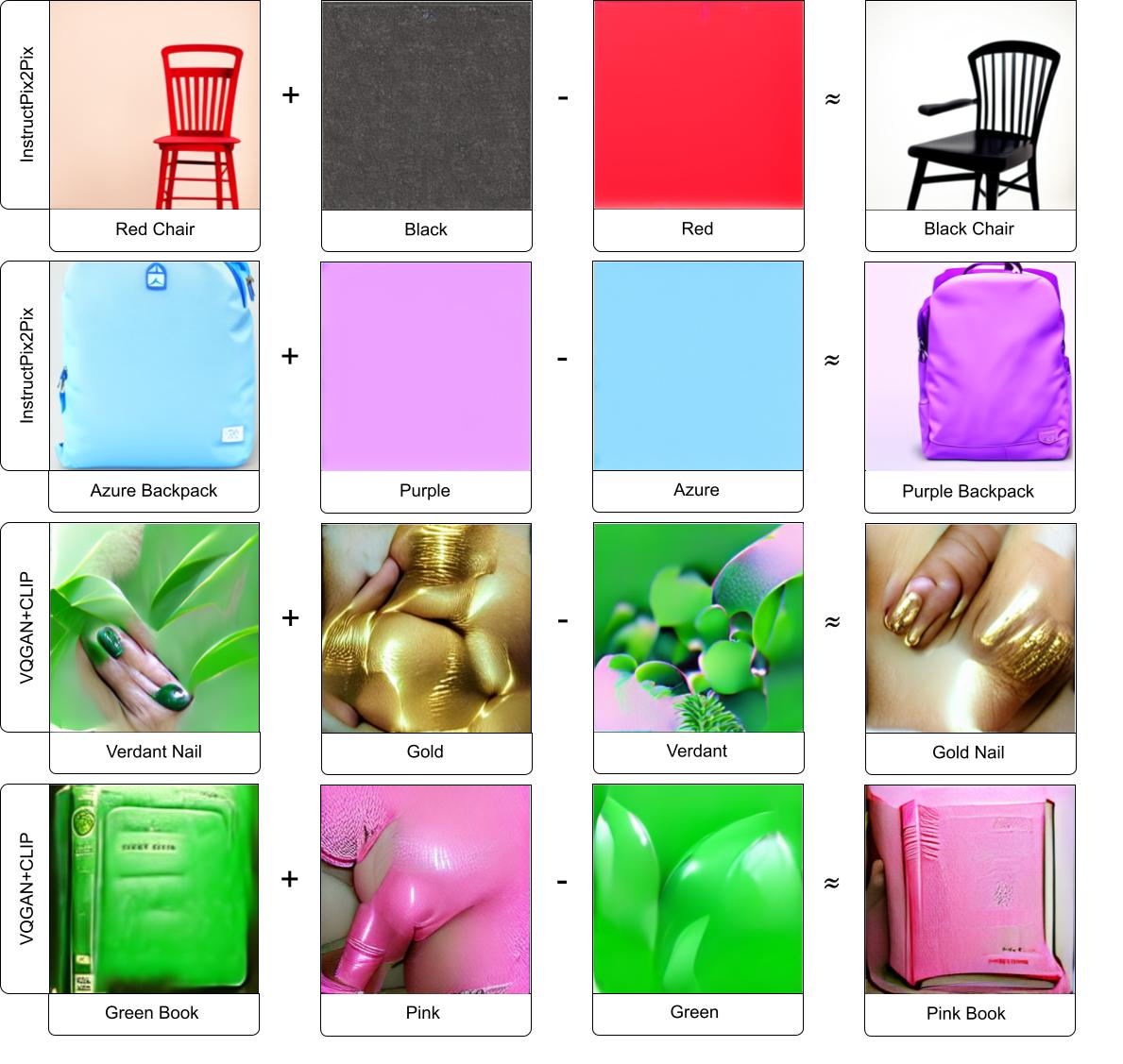}
    \caption{Color examples}
    \label{fig:color-label}
\end{figure}

\begin{figure}
    \centering
    \includegraphics[scale = 0.3]{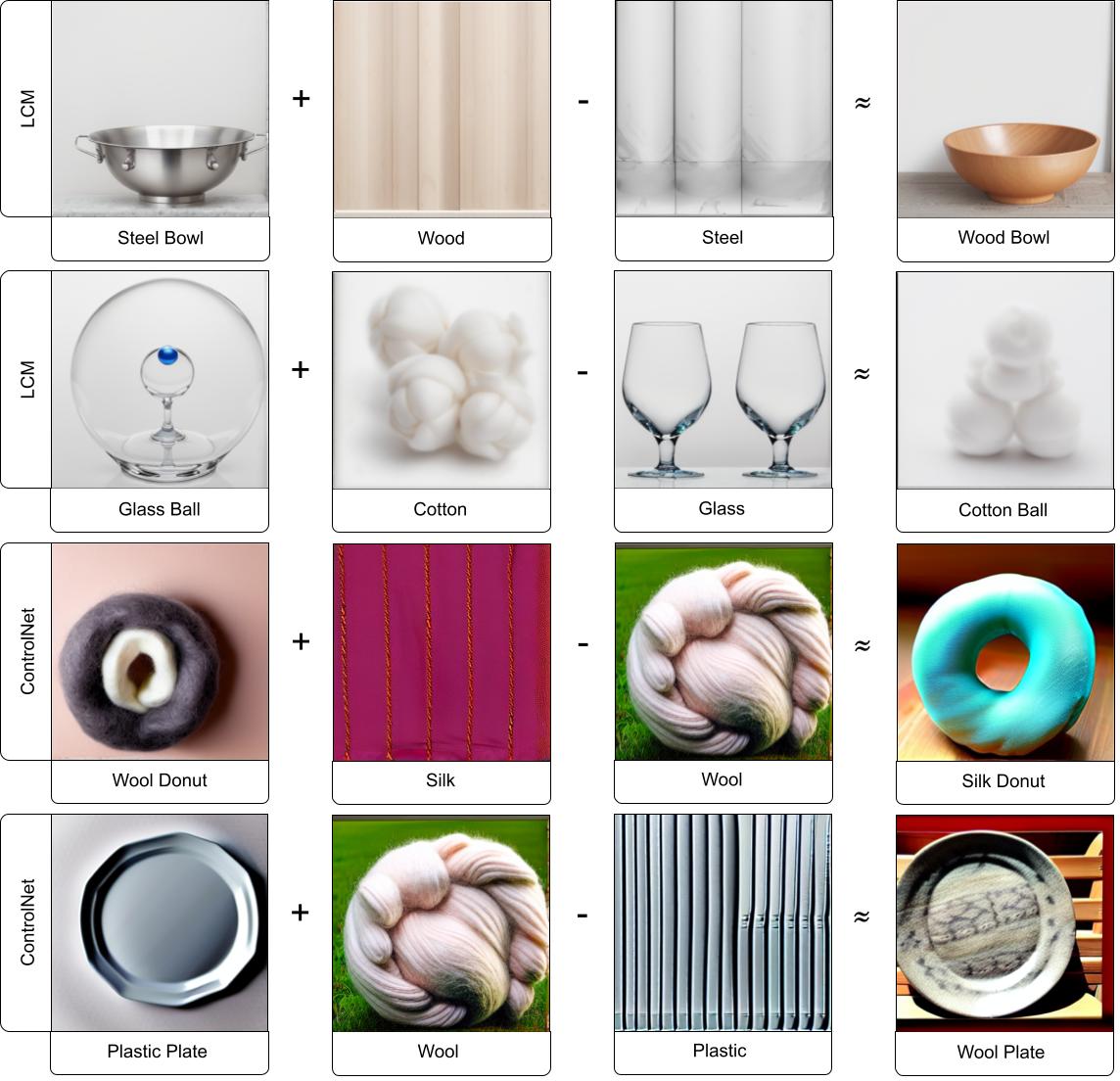}
    \caption{Texture examples}
    \label{fig:texture-label}
\end{figure}

\begin{figure}
    \centering
    \includegraphics[scale = 0.3]{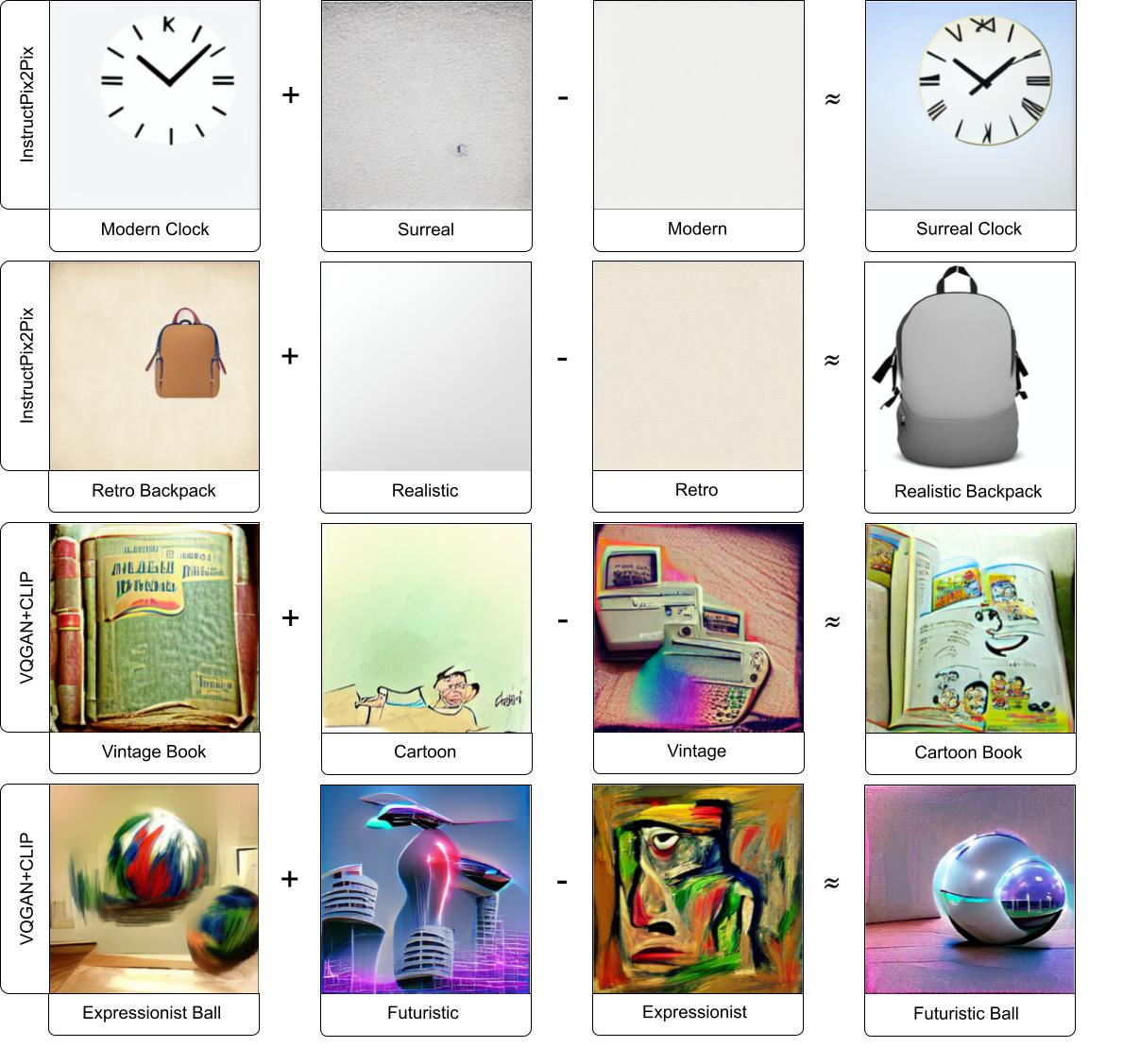}
    \caption{Style examples}
    \label{fig:style-label}
\end{figure}

\begin{figure}
    \centering
    \includegraphics[scale = 0.3]{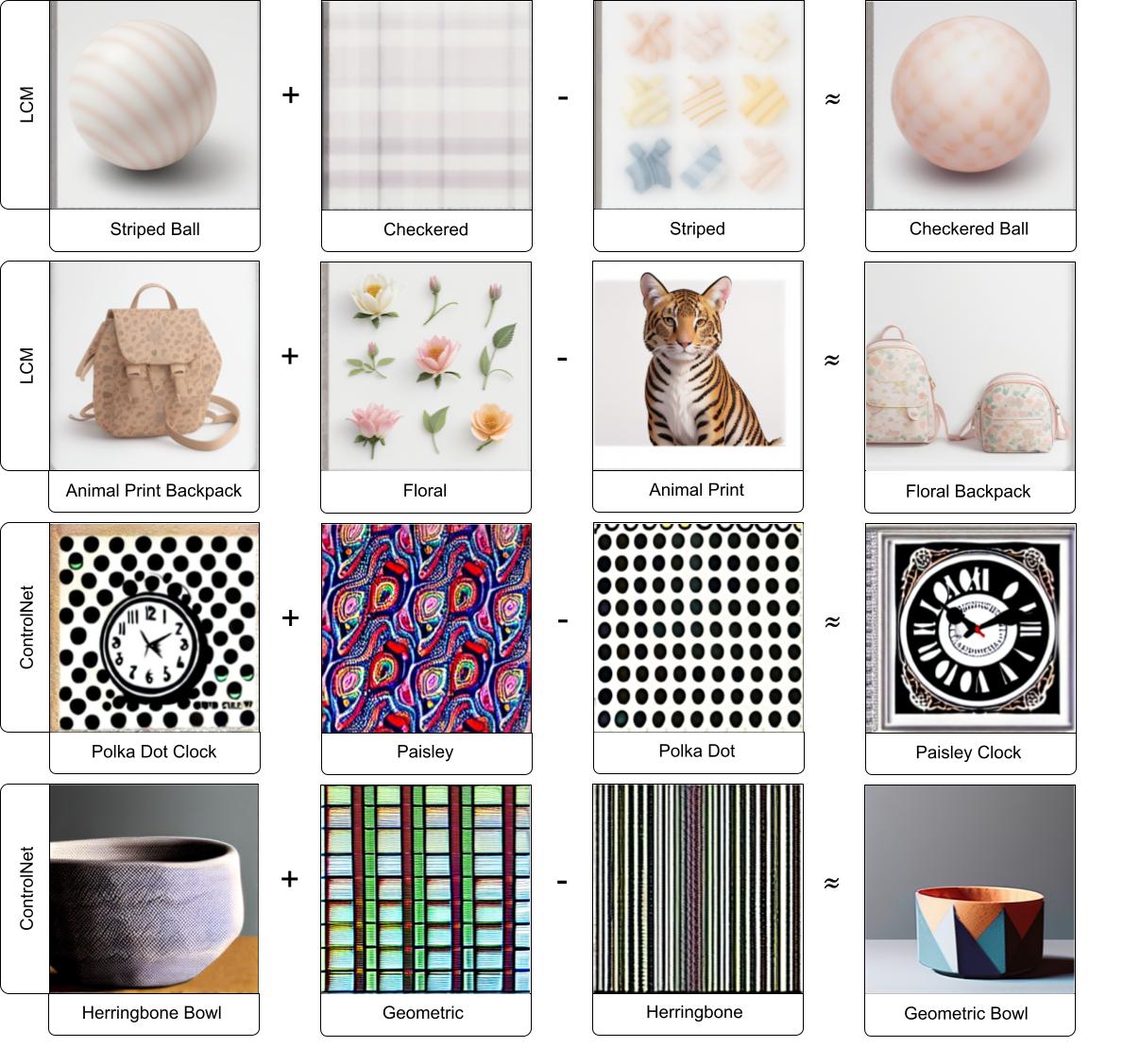}
    \caption{Pattern examples}
    \label{fig:pattern-label}
\end{figure}

We provide many examples for the four different attributes in  \cref{fig:color-label,fig:texture-label,fig:style-label,fig:pattern-label}. Each row corresponds to the first part of the disentanglement evaluation, tested in one of the models (left), and each image is the result of using the model with the corresponding prompt in the white image of reference.

\end{document}